%% file: paper.tex
\pdfoutput=1
\documentclass[10pt,twocolumn,letterpaper]{article}

\usepackage{wacv}
\usepackage{times}
\usepackage{epsfig}
\usepackage{graphicx}
\usepackage{amsmath}
\usepackage{amssymb}

\usepackage[utf8]{inputenc} 
\usepackage[T1]{fontenc}    
\usepackage{hyperref}
\usepackage{url}
\usepackage{booktabs}       
\usepackage{amsfonts}       
\usepackage{nicefrac}       
\usepackage{microtype}      
\usepackage{multirow,graphicx,paralist}
\usepackage{float}
\usepackage{wrapfig,lipsum}
\usepackage{url}
\usepackage{dsfont,amssymb}
\usepackage{authblk}

\newcommand{\rot}{\rotatebox{90}}

\newcommand\Tstrut{\rule{0pt}{2.6ex}}

\wacvfinalcopy 

\begin{document}

\title{Label Refinement Network for Coarse-to-Fine Semantic Segmentation}


\author[1]{Md Amirul Islam}
\author[2]{Shujon Naha}
\author[1]{Mrigank Rochan}
\author[1]{Neil Bruce}
\author[1]{Yang Wang}
\affil[1]{University of Manitoba, Winnipeg, MB }
\affil[2]{Indiana University, Bloomington, IN \vspace{-2ex}} 

\maketitle

\begin{abstract}
  We consider the problem of semantic image segmentation using deep convolutional neural networks. We propose a novel network architecture called the \emph{label refinement network} that predicts segmentation labels in a coarse-to-fine fashion at several resolutions. The segmentation labels at a coarse resolution are used together with convolutional features to obtain finer resolution segmentation labels. We define loss functions at several stages in the network to provide supervisions at different stages. Our experimental results on several standard datasets demonstrate that the proposed model provides an effective way of producing pixel-wise dense image labeling.
\end{abstract}

\input intro.tex

\input related.tex

\input approach.tex
\input exp1.tex
\input conclude.tex

{\small
\bibliographystyle{ieee}
\bibliography{wang}
}

\end{document}

%% file: intro.tex
\section{Introduction}
\begin{figure*}[t]
    \centering
    \includegraphics[width=\textwidth]{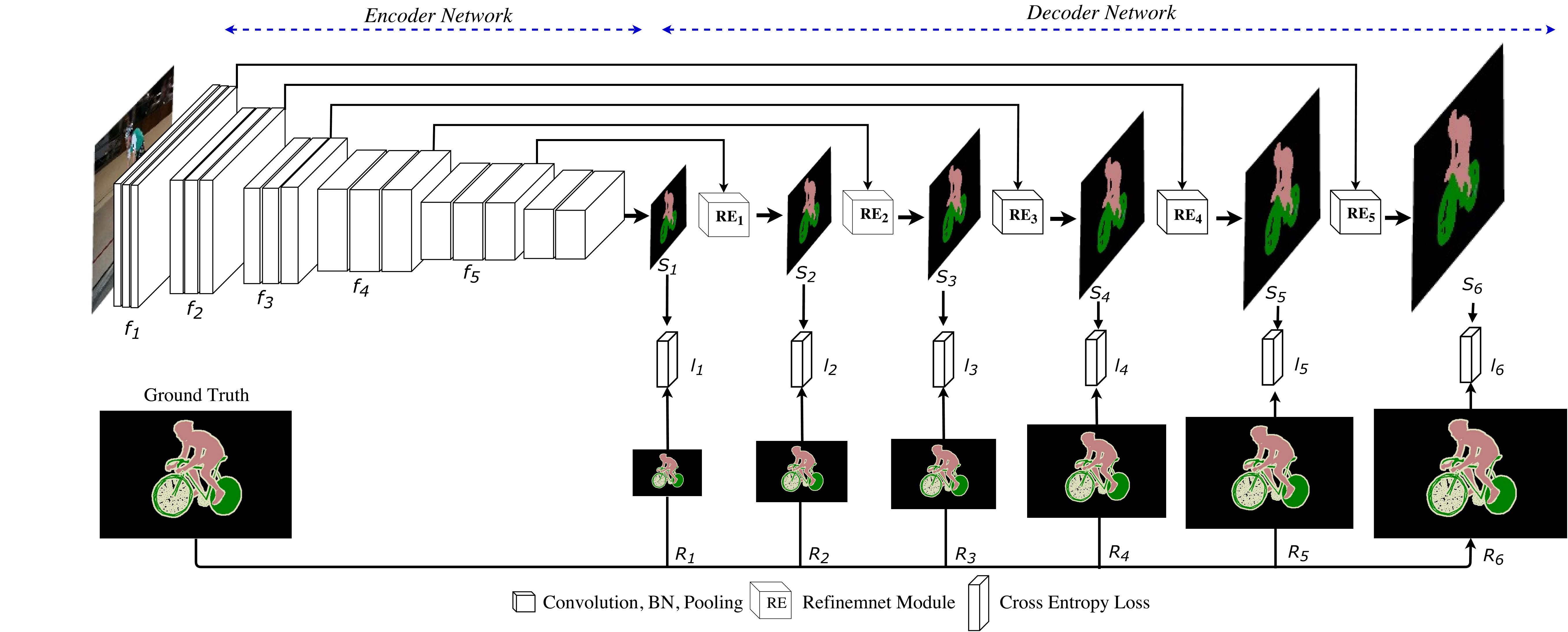}
    \caption{Overall architecture of the Label Refinement Network (LRN). LRN is based on encoder-decoder framework. The encoder network produces a sequence of feature maps with decreasing spatial dimensions. The decoder network produces label maps with increasing spatial dimensions. A larger label map is obtained by combining the previous (smaller) label map and the corresponding convolutional features from a layer in the encoder network indicated by the solid line. We use down-sampled ground-truth label maps to provide the supervision at each stage of the decoder network.}
 \label{fig:our_archi}
\end{figure*}
We consider the problem of semantic image segmentation, where the goal is to densely label each pixel in an image according to the object class that it belongs to. We propose a convolutional neural network architecture called the \emph{label refinement network}~(LRN) that performs semantic segmentation in a coarse-to-fine fashion.

Deep convolutional neural networks (CNNs) have been successfully applied to a wide variety of visual recognition problems, such as image classification~\cite{krizhevsky12_nips}, object detection~\cite{ren15_nips}, action recognition~\cite{tran15_iccv}, etc. CNNs extract deep feature hierarchies using alternating layers of operations, such as convolution, pooling, etc. Features from the top layers of CNNs tend to be invariant to ``nuisance factors'' such as pose, illumination, small translations, etc. The invariance properties of these features make them particularly useful for vision tasks such as whole image classification. However, these features are not well suited for other vision tasks~(e.g. semantic segmentation) that require precise pixel-wise information.

There have been some recent efforts~\cite{badrinarayanan15_arxiv,chen15_iclr,farabet13_pami,hariharan15_cvpr,long15_cvpr,mostajabi15_cvpr,noh15_iccv} on adapting CNNs for semantic segmentation. Some of these approaches~(e.g.~\cite{hariharan15_cvpr,long15_cvpr,mostajabi15_cvpr}) are based on combining the convolutional features from multiple layers in a CNN. Another popular approach~(e.g.~\cite{badrinarayanan15_arxiv,noh15_iccv} is to use upsampling (also known as deconvolution) to enlarge the spatial dimensions of the feature map at the top layer of a CNN, e.g. to the same spatial dimensions as the original image, then predict the pixel-wise labels from the enlarged feature map. In both the cases, the final full-sized semantic segmentation result is obtained in a ``single shot'' at the very end of the network architecture.

In this paper, we introduce a novel CNN-based architecture for semantic segmentation. Different from previous approaches, our proposed model predicts semantic labels at several different resolutions in a coarse-to-fine fashion. We use resized ground-truth segmentation labels as the supervision at each resolution level. The segmentation labels at a coarse scale are combined with convolutional features to produce segmentation labels at a finer scale. See Fig.~\ref{fig:our_archi} for an illustration of our model architecture.

We make threefold contributions in this paper which are as follows:
\begin{itemize}
 \item We introduce a new perspective on the semantic segmentation (or more generally, pixel-wise labeling) problem. Instead of predicting the final segmentation result in a single shot, we propose to solve the problem in a coarse-to-fine fashion by first predicting a coarse labeling, then progressively refine the result to get the finer scale results. 
 \item We propose an end-to-end CNN architecture to learn to predict the segmentation labels at multiple resolutions. Unlike most of the previous methods that only have supervision at the end of their network, our model has supervision at multiple resolutions in the network. Although we focus on semantic image segmentation in this paper, our network architecture is general enough to be used for any pixel-wise labeling task. 
 \item We perform extensive experiments on several standard datasets to demonstrate the effectiveness of our proposed model.
\end{itemize}

The rest of the paper is organized as follows. Section~\ref{related} presents related work. Section~\ref{back} discusses the essential background of encoder-decoder architecture. Section~\ref{app} introduces our label refinement network. Section~\ref{exp} describes the experiment details and also presents some analysis. Finally, we conclude the paper in Section~\ref{conclude}.	 





%% file: related.tex
\section{Related Work}\label{related}
CNNs have shown tremendous success in various visual recognition tasks, e.g. image classification~\cite{krizhevsky12_nips}, object detection~\cite{ren15_nips}, action recognition~\cite{tran15_iccv}. Recently, there has been work~\cite{badrinarayanan15_arxiv,chen15_iclr,farabet13_pami,hariharan15_cvpr,long15_cvpr,mostajabi15_cvpr,noh15_iccv} on adapting CNNs for pixel-wise image labeling problems such as semantic segmentation.

Directly related to our work is the idea of multi-scale processing in computer vision. Early work on Laplacian pyramids~\cite{burt83} is an example of capturing image structures at multiple scales. Eigen et al.~\cite{eigen15_iccv} use a multi-scale CNN for predicting depths, surface normals and semantic labels from a single image. Denton et al. \cite{denton15_nips} use coarse-to-fine idea in the context of image generation. Honari et al.~\cite{honari16_cvpr} combine coarse-to-fine features in CNNs for facial keypoint localization.

Similar to our proposed model, some papers have also used the idea of defining loss functions at multiple stages in a CNN to provide more supervisions in learning. The Inception model~\cite{szegedy15_cvpr} use auxiliary classifiers at the lower stages of the network to encourage the network to learn discriminative features. Lee et al.~\cite{lee15_aistats} propose a deeply-supervised network for image classification. Xie et al.~\cite{xie15_iccv} use a similar idea for edge detection.

%% file: approach.tex
\begin{figure*}
\centering
 \includegraphics[width=0.99\textwidth]{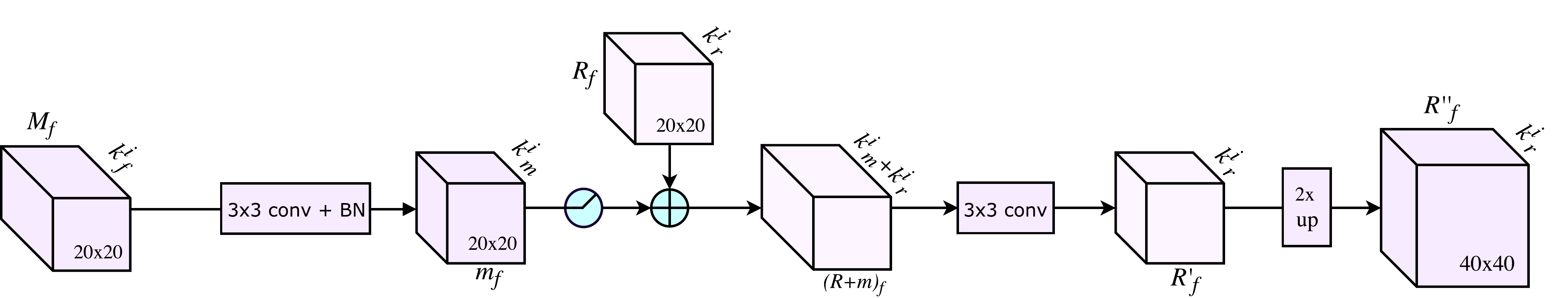}
 \caption{Detailed illustration of the refinement module (RE). The refinement module (see RE in Fig.~\ref{fig:our_archi}) is used after each segmentation map in the decoder network. Here we unfold and illustrate RE for $i^{th}$ $(i=1,2,..,5)$ stage of the decoder. RE takes two inputs. The first input is the coarse label map $R_f$ with channel dimension $k^i_r$ generated at a decoding stage, and the second input is the skip feature map $M_f$ with channel dimension $k^i_f$ from the corresponding convolution layer (we use the last convolution layer of each stage ($f_{c}$ in Fig.~\ref{fig:our_archi}) where $c=1,2,..,5$)) of the encoder. $R_f$ and $M_f$ have the same spatial dimension but different channel dimensions (typically $k^i_m \gg k^i_r$). In order to combine $R_f$ and $M_f$, we apply a 3x3 convolution on $M_f$ followed by a batch normalization and ReLU operation. These operations results in a skip feature map $m_f$ (with channel dimension $k^i_m$) that matches in terms of the spatial dimension and channel dimension of the coarse label map $R_f$. We then apply a $3\times 3$ convolution to the combined feature map $(R$+$m)_f$ to produce a feature map $R^\prime_f$ whose channel dimension is same as of coarse label feature map $R_f$. $R^\prime_f$ is the $i^{th}$ stage prediction map. Then we upsample the prediction map $R^\prime_f$ by a factor of 2 ($R^{\prime\prime}_f$ in the figure which is the segmentation map $s_k(I)$ in Fig.~\ref{fig:our_archi}) and feed it to the next stage $(i+1)^{th}$ of the decoder. Note that convolution and upsampling layers are shown in rectangles, whereas feature maps and concatenation operations are represented by 3D boxes and circles respectively. In summary, the refinement module is composed of convolution, batch normalization, ReLU, concatenation, and bilinear upsampling operations.}
 \label{fig:refinement}
\end{figure*}
\section{Background: Encoder-Decoder Network for Semantic Segmentation}\label{back}
In recent years, convolutional neural networks have been driving a lot of success in computer vision. Following the early success of CNNs on image classification~\cite{krizhevsky12_nips}, there has been a line of work~\cite{badrinarayanan15_arxiv,long15_cvpr,noh15_iccv} on adapting CNNs for pixel-wise dense image labeling problems such as semantic image segmentation. Most of the CNN-based semantic segmentation algorithms can be characterized in terms of a generic encoder-decoder architecture~\cite{badrinarayanan15_arxiv,noh15_iccv}. The encoder network is usually a convolutional network that extracts features from an input image. Similar to the CNNs used for image classification, the encoder network typically consists of alternating layers of convolution, subsampling (e.g. via max-pooling), non-linear rectification, etc.

The subsampling layers in CNNs allow the networks to extract high-level features that are translation invariant, which are crucial for image classification. However, they reduce the spatial extent of the feature map at each layer in the network. Let $I\in\mathbb{R}^{h\times w\times d}$ be the input image~(where $h$, $w$ are spatial dimensions $d$ is the color channel dimension) and $f(I)\in\mathbb{R}^{h'\times w'\times d'}$ be the feature map volume at the end of the encoder network. The feature map $f(I)$ has smaller spatial dimensions than the original image, i.e. $h'<h$ and $w'<w$. In order to produce a full-sized segmentation result as the output, an associated decoder network is applied to the output of the encoder network to produce the result that has spatial size of the original image. The fundamental differences between our work and various research contributions in prior work mainly lie in choices of the structure and composition of the decoder network. The decoder in SegNet~\cite{badrinarayanan15_arxiv} progressively enlarges the feature map using an upsampling technique without learnable parameters. The decoder in DeconvNet~\cite{noh15_iccv} is similar, but has learnable parameters. FCN~\cite{long15_cvpr} uses a single layer interpolation for deconvolution.

At the end of the decoder network, the full-sized feature map is mapped to predict the class label for each pixel, and a loss function is defined to compare the predicted labels with the ground-truth. It is important to note that the network makes the prediction only at the end of the decoder network. As a result, the supervision only happens at the last layer of the entire architecture, with which the loss function is associated.

\section{Label Refinement Network}\label{app}
In this section, we propose a novel network architecture called the \emph{Label Refinement Network~(LRN)} for dense image labeling problems. An overview of the LRN architecture is shown in Fig.~\ref{fig:our_archi}. Similar to prior work~\cite{badrinarayanan15_arxiv,long15_cvpr,noh15_iccv}, LRN also uses an encoder-decoder framework. The encoder network of LRN is similar to that of SegNet~\cite{badrinarayanan15_arxiv} which is based on the VGG16 network~\cite{simonyan15_iclr}. The novelty of LRN lies in the decoder network. Instead of making the prediction at the end of the entire architecture, the decoder network in LRN makes predictions in a coarse-to-fine fashion at several stages. In addition, LRN introduces supervisions early in the network by adding loss functions at multiple stages (not just at the last layer) in the decoder network. 

The LRN architecture is motivated by the following observations: Due to the convolution and subsampling operations, the feature map $f(I)$ obtained at the end of the encoder network mainly contains high-level information about the image~(e.g. objects). Spatially precise information is lost in the encoder network, and therefore $f(I)$ cannot be used directly to recover a full-sized semantic segmentation which requires pixel-precise information. However, $f(I)$ contains enough information to produce a \emph{coarse} semantic segmentation. In particular, we can use $f(I)$ to produce a segmentation map $s(I)$ of spatial dimensions $h'\times w'$, which is smaller than the original image dimensions $h\times w$. Our decoder network then progressively refines the segmentation map $s(I)$. Let $s_k(I)$ be the $k$-th segmentation map (left to right in Fig.~\ref{fig:our_archi}). If $k>k'$, the segmentation map $s_k(I)$ will have a larger spatial dimensions than $s_{k'}(I)$. Note that one can interpret $s_k(I)$ as the feature map in the decoder network in most work~(e.g. \cite{badrinarayanan15_arxiv,long15_cvpr,noh15_iccv}). However, our model enforces the channel dimension of $s_k(I)$ to be the same as the number of class labels, so $s_k(I)$ can be considered as a (soft) label map.

The network architecture in Fig.~\ref{fig:our_archi} has 5 convolution layers in the encoder. We use $f_{k}(I)\in\mathbb{R}^{h_k\times w_k \times d_k}$ ($k=1,2,...,5$) to denote the feature map after the $k$-th convolution layer. The decoder network generates 6 label maps $s_{k}$ ($k=1,2,...,6$). After the last convolution layer of the encoder network, we use a $3\times 3$ convolution layer to convert the convolution feature map $f_5(I)\in\mathbb{R}^{h_5\times w_5\times d_5}$ to $s_{1}(I)\in\mathbb{R}^{h_5\times w_5\times C}$, where $C$ is the number of class labels. We then define a loss function on $s_{1}(I)$ as follows. Let $Y\in\mathbb{R}^{h\times w \times C}$ be the ground-truth segmentation map, where the label on each pixel is represented as a $C$-dimensional vector using the one-shot representation. We use $R_{1}(Y)\in\mathbb{R}^{h_5\times w_5\times C_5}$ to denote the segmentation map obtained by resizing $Y$ to have the same spatial dimensions as $s_{1}(I)$. We can then define a loss function $\ell_1$ (cross entropy loss is used) to measure the difference between the resized ground-truth $R_{1}(Y)$ and the coarse prediction $s_{1}(I)$ (after softmax). In other words, these operations can be written as:
\begin{eqnarray}
s_{1}(I)=\textrm{conv}_{3\times 3}\big(f_5(I)\big) \nonumber \\
\qquad \ell_{1}=\textrm{Loss}\Big(R_{1}(Y),\textrm{softmax}\big(s_{1}(I)\big)\Big)
\end{eqnarray}
where $\textrm{conv}_{3\times 3}(\cdot)$ denotes the $3\times 3$ convolution and $\textrm{Loss}(\cdot)$ denotes the cross entropy loss function.

Now we explain how to get the subsequent segmentation map $s_{k}(I)$ ($k>1$) with larger spatial dimensions. One simple solution is to upsample the previous segmentation map $s_{k-1}(I)$. However, this upsampled segmentation map will be very coarse. In order to derive a more precise segmentation, we use the idea of skip-connections \cite{hariharan15_cvpr,long15_cvpr}. The basic idea is to make use of the outputs from one of the convolutional feature layers in the encoder network. Since the convolutional feature layer contains more precise spatial information, we can combine it with the upsampled segmentation map to obtain a refined segmentation map~(see Fig~\ref{fig:refinement} for detailed illustration of our refinement module). In our model, we simply concatenate the outputs from the skip layer with the upsampled decoder layer to create a larger feature map, then use $3\times 3$ convolution to convert the channel dimension to $C$. For example, the label map $s_{2}(I)$ is obtained from upsampled $s_{1}(I)$ and the convolutional feature map $f_{5}(I)$ (see Fig.~\ref{fig:our_archi} for details on these skip connections). We can then define a loss function on this (larger) segmentation map by comparing $s_{k}(I)$ with the resized ground-truth segmentation map $R_{k}(Y)$ of the corresponding size. These operations can be summarized as follows:
\begin{eqnarray}
  s_{k}(I)=\textrm{conv}_{3\times 3}\Big(\textrm{concat}\Big(\textrm{upsample}\big(s_{k}(I)\big), f_{7-k}(I)\Big)\Big)\\
  \ell_{k}=\textrm{Loss}\Big(R_{k}(Y),\textrm{softmax}\big(s_{k}(I)\big)\Big), \ \ \textrm{where } k=2,..,6
\end{eqnarray}
Our model is trained using back-propagation to optimize the summation of all the losses $\sum_{k=1}^{6} \ell_{k}$.

\begin{figure}
\center
 \includegraphics[width=2.5in, height=4.5in]{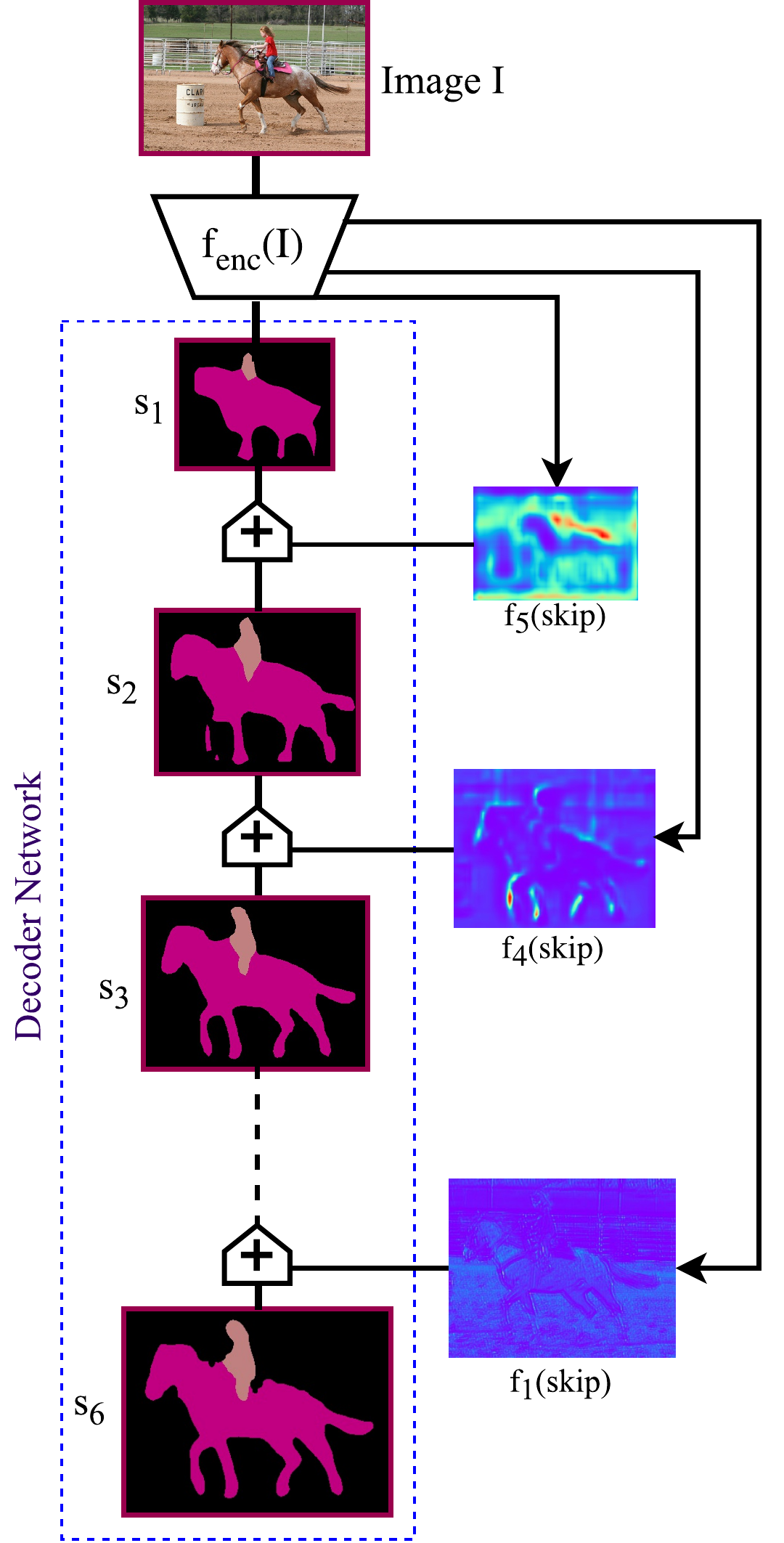}
 \caption{Visualization of the refinement process in the network. Coarse segmentation map ($s_1$) produced by the encoder network $f_{enc}$ is fed to the decoder network (shown in dotted rectangular box). The decoder network then progressively refines the obtained segmentation map using the refinement module (RE) (see Fig.~\ref{fig:refinement}) to recover the image details step by step (see $s_1,s_2,s_3,...s_6$ in the figure). Basically, the refinement process is integrating the finer feature maps (activations shown as heat maps on the right) from the corresponding convolution layers with the segmentation maps. As a result, the segmentation map gets progressively refined throughout the decoder network.}
 \label{fig:skip_visualize}
\end{figure}

Fig.~\ref{fig:skip_visualize} shows the advantage of progressively refinement of the segmentation map $s(I)$ in the decoder network. The finer feature maps from the convolution layers help to recover image details, and therefore the segmentation map produced at each stage of the decoder network becomes more accurate.

%% file: exp1.tex
\section{Experiments}\label{exp}
Section~\ref{sec:implementation} explains some implementation details and several baseline methods that we compare with. We then demonstrate experimental results on three challenging benchmark datasets: the PASCAL VOC 2012 dataset~(Sec.~\ref{sec:pascal}), the Cambridge Driving Labeled Video (CamVid) dataset~(Sec.~\ref{sec:camvid}), and the SUN RGB-D dataset~(Sec.~\ref{sec:sun}).
\begin{table*}[ht!]
\resizebox{\textwidth}{!}{%
    \centering
    \begin{tabular}{r|cccccccccccccccccccc|c}
        method&\multicolumn{1}{c}{\rot{aero}}&\multicolumn{1}{c}{\rot{bike}}&\multicolumn{1}{c}{\rot{bird}}&\multicolumn{1}{c}{\rot{boat}}&\multicolumn{1}{c}{\rot{bottle}}&\multicolumn{1}{c}{\rot{bus}}&\multicolumn{1}{c}{\rot{car}}&\multicolumn{1}{c}{\rot{cat}}&\multicolumn{1}{c}{\rot{chair}}&\multicolumn{1}{c}{\rot{cow}}&\multicolumn{1}{c}{\rot{table}}&\multicolumn{1}{c}{\rot{dog}}&\multicolumn{1}{c}{\rot{horse}}&\multicolumn{1}{c}{\rot{mbike}}&\multicolumn{1}{c}{\rot{person}}&\multicolumn{1}{c}{\rot{plant}}&\multicolumn{1}{c}{\rot{sheep}}&\multicolumn{1}{c}{\rot{sofa}}&\multicolumn{1}{c}{\rot{train}}&\rot{tv}&\rot{mean IoU}\\
       \hline
       \hline
       FCN-8s \cite{long15_cvpr}&76.8& 34.2& 68.9& 49.4& {\bf 60.3}& 75.3& 74.7& 77.6& 21.4 &62.5& 46.8& {\bf 71.8}& 63.9& 76.5& 73.9& 45.2& {\bf 72.4}& 37.4 &70.9& 55.1&62.2\\



       SegNet \cite{badrinarayanan15_arxiv}&74.5& 30.6& 61.4& {\bf 50.8}& 49.8& 76.2& 64.3&69.7& {\bf 23.8}& 60.8& {\bf 54.7}&62.0& 66.4& 70.2& 74.1& 37.5& 63.7& 40.6& 67.8& 53.0&59.1\\
       SegNet+ DS&70.0& 35.7& 72.8& 42.7& 53.0& 75.6& 71.6& 72.8& 23.2& {\bf 69.5}& 46.8& 64.6& 68.8& 78.9& 76.0& 41.9& 65.7& {\bf 44.1}& 65.7& 53.7& 61.1\Tstrut\\
       \textbf{LRN}&{\bf 79.3}&{\bf 37.5}&{\bf 79.7}&47.7&58.3&{\bf 76.5}&{\bf 76.1}&{\bf 78.5}&21.9&67.7 &47.6&71.2&{\bf 69.1}&{\bf 82.1}&{\bf 77.5}&{\bf 46.8} &70.1&40.3 &{\bf 71.5}&{\bf 57.4}&{\bf 64.2}  \Tstrut\\
       \hline
    \end{tabular}}
\caption{Quantitative results on the PASCAL VOC 2012 test set. We report the IoU score for each class and the mean IoU score.}
  \label{tab:quant_pascal}

\end{table*} 
\begin{figure*}[ht!]
    \begin{center}
    \setlength\tabcolsep{0.7pt}
    \def\arraystretch{0.5}
    \resizebox{\textwidth}{!}{
 \begin{tabular}{*{4}{c }}
    \includegraphics[width=0.2\textwidth]{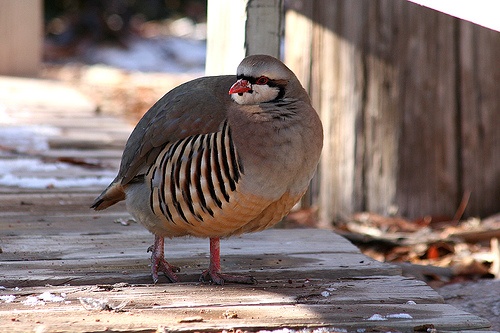}
   &
   \includegraphics[width=0.2\textwidth]{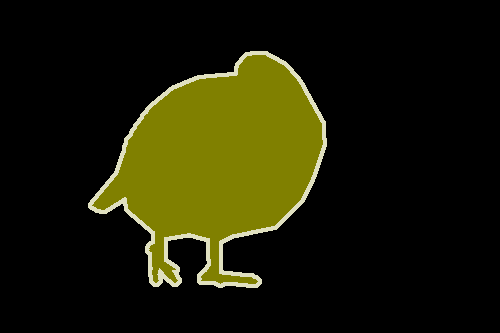}
   &
  \includegraphics[width=0.2\textwidth]{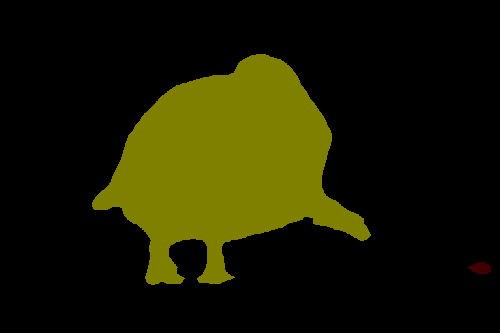}
  &
  {\includegraphics[width=0.2\textwidth]{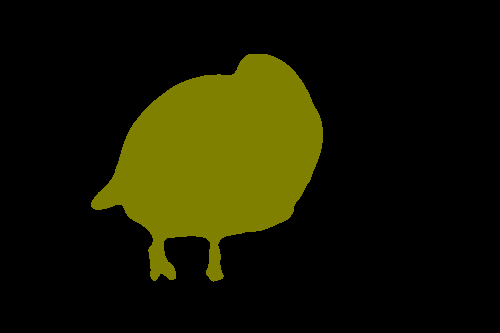}} \\
   \includegraphics[width=0.2\textwidth]{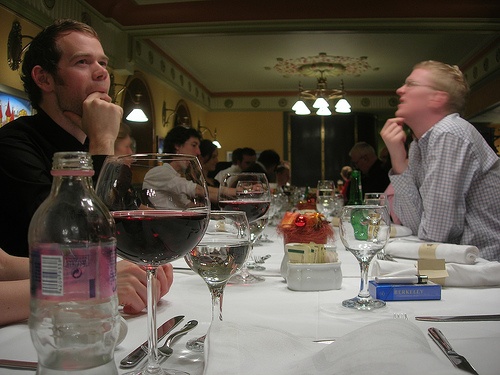}
   &
   \includegraphics[width=0.2\textwidth]{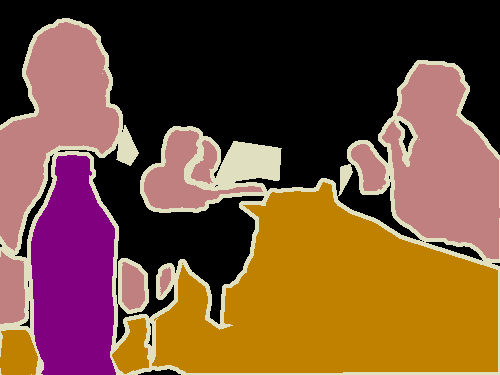}
   &
  \includegraphics[width=0.2\textwidth]{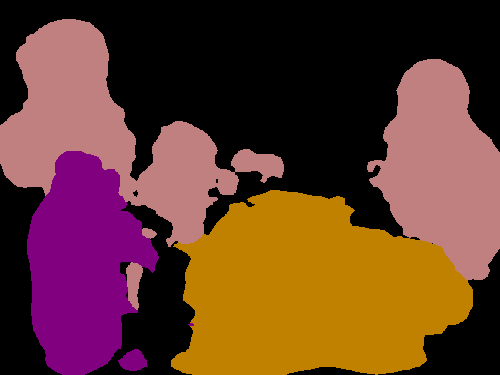}
  &
  {\includegraphics[width=0.2\textwidth]{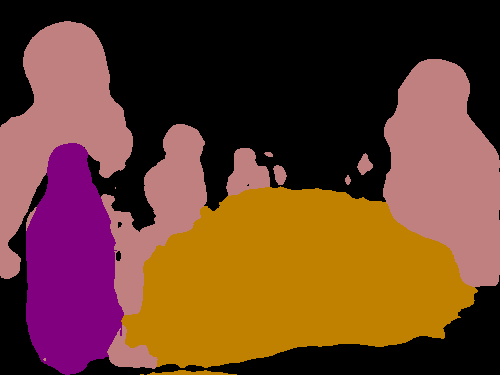}} \\
   \includegraphics[width=0.2\textwidth]{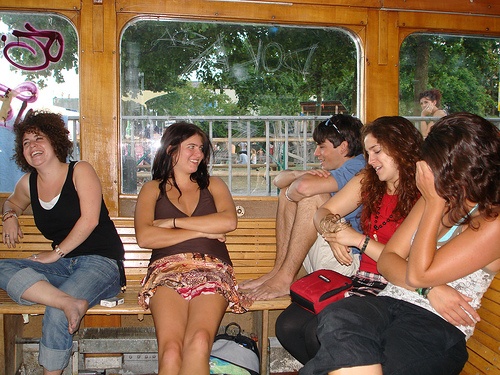}
   &
   \includegraphics[width=0.2\textwidth]{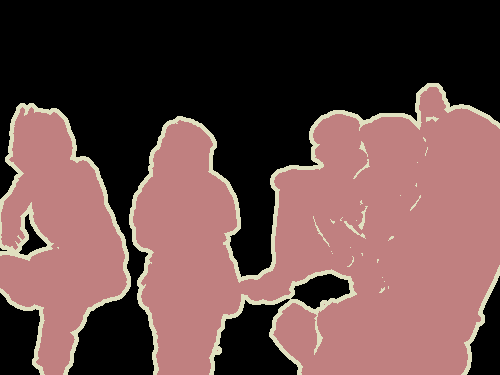}
   &
  \includegraphics[width=0.2\textwidth]{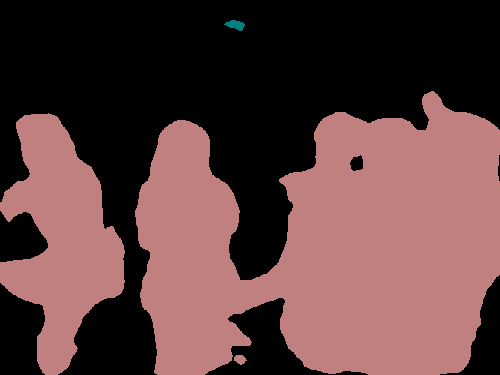}
  &
  {\includegraphics[width=0.2\textwidth]{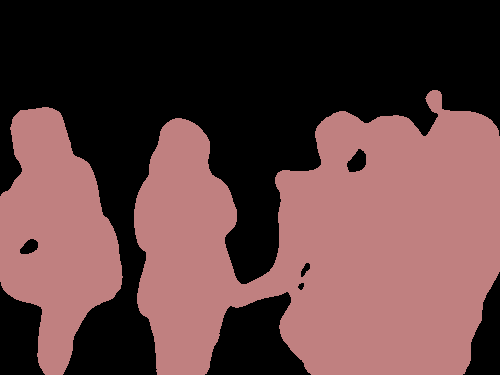}} \\
  \includegraphics[width=0.2\textwidth]{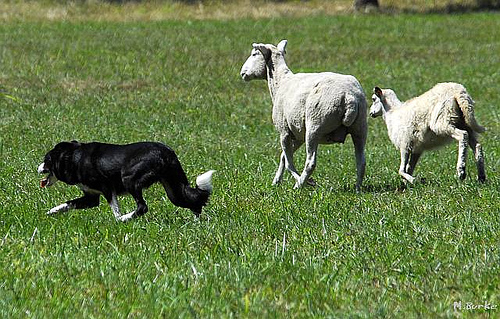}
   &
   \includegraphics[width=0.2\textwidth]{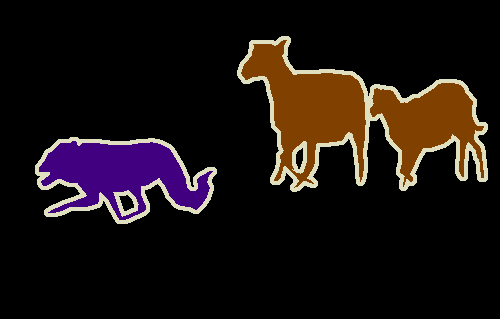}
   &
  \includegraphics[width=0.2\textwidth]{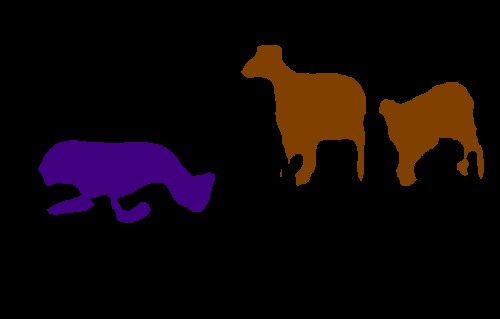}
  &
  {\includegraphics[width=0.2\textwidth]{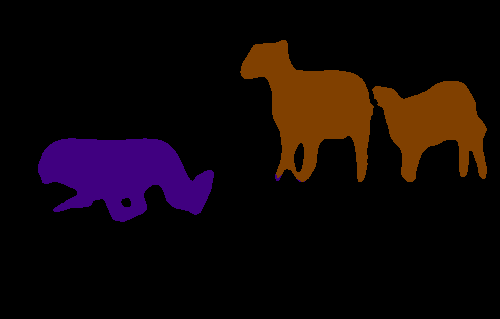}} \\
  \includegraphics[width=0.2\textwidth]{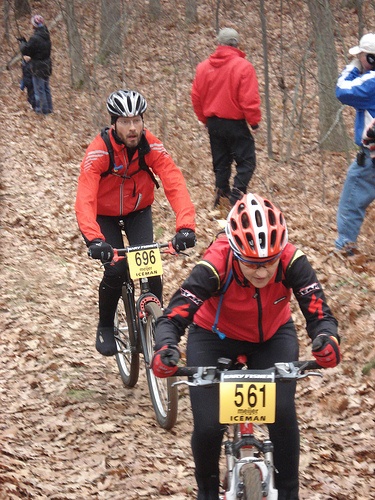}
   &
   \includegraphics[width=0.2\textwidth]{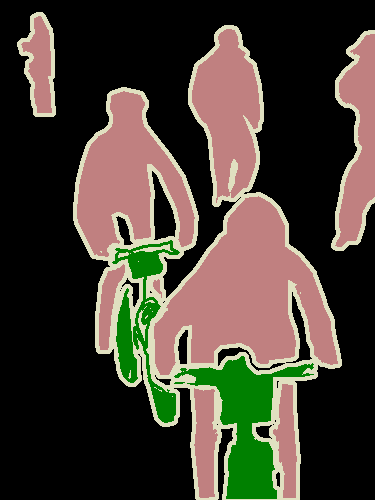}
   &
  \includegraphics[width=0.2\textwidth]{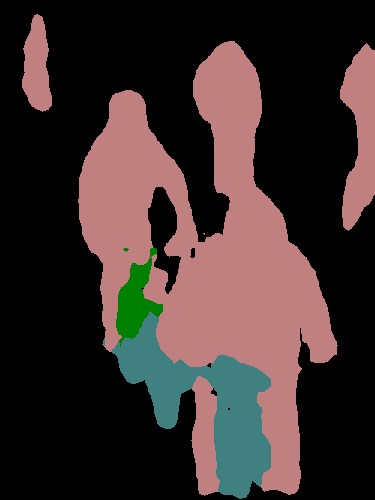}
  &
  {\includegraphics[width=0.2\textwidth]{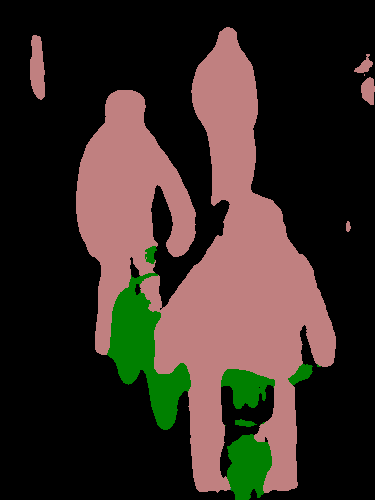}} \\
                  image & ground-truth & FCN-8s \cite{long15_cvpr} & LRN \Tstrut\\
  \end{tabular}
  \begin{tabular}{*{4}{c }}
   \includegraphics[width=0.2\textwidth]{images/pascal/original/2007_007235.jpg}
   &
   \includegraphics[width=0.2\textwidth]{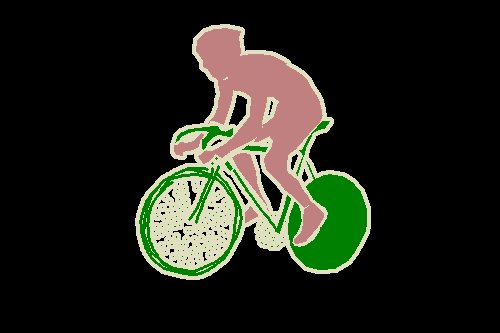}
   &
  \includegraphics[width=0.2\textwidth]{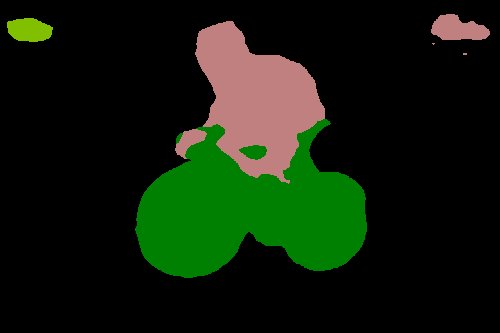}
  &
  {\includegraphics[width=0.2\textwidth]{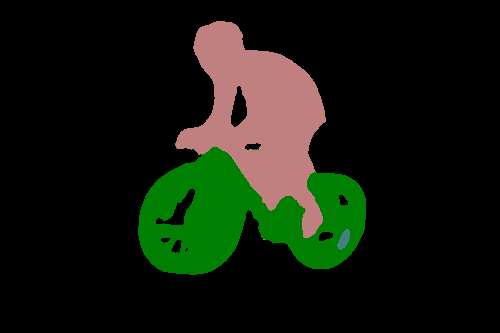}} \\
  \includegraphics[width=0.2\textwidth]{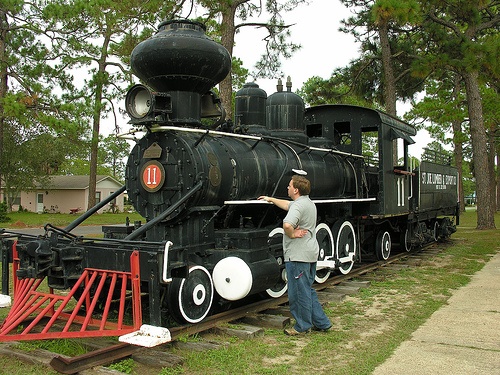}
   &
   \includegraphics[width=0.2\textwidth]{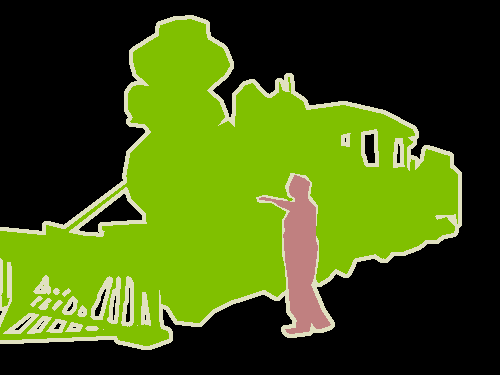}
   &
  \includegraphics[width=0.2\textwidth]{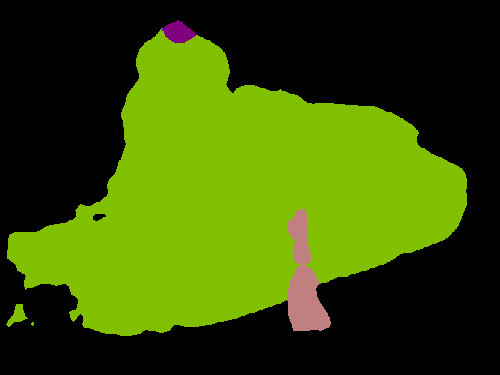}
  &
  {\includegraphics[width=0.2\textwidth]{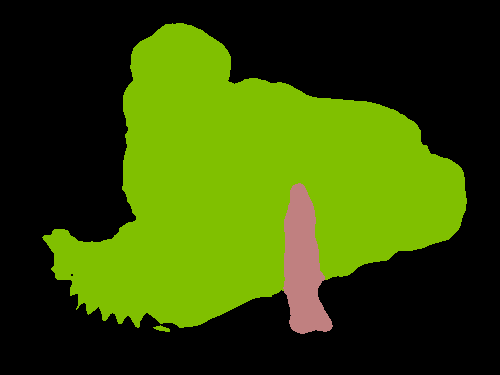}} \\
   \includegraphics[width=0.2\textwidth]{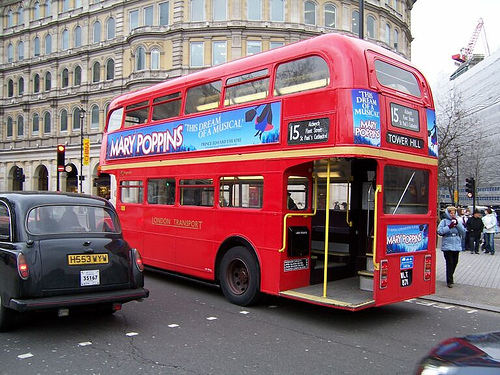}
   &
   \includegraphics[width=0.2\textwidth]{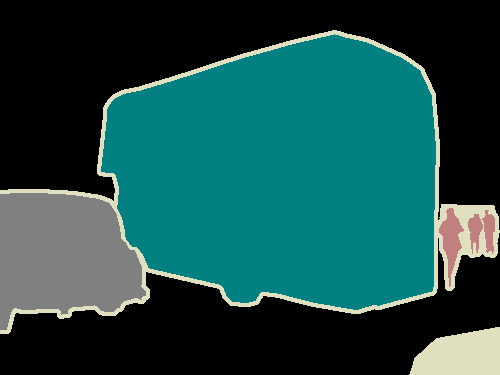}
   &
  \includegraphics[width=0.2\textwidth]{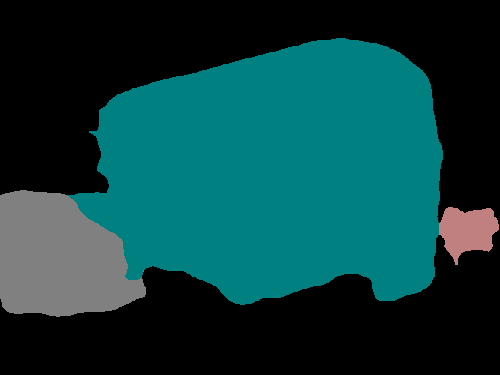}
  &
  {\includegraphics[width=0.2\textwidth]{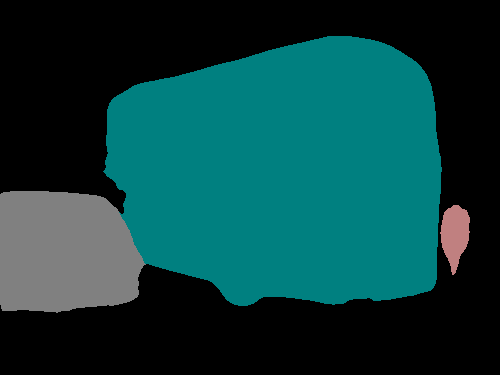}} \\
   \includegraphics[width=0.2\textwidth]{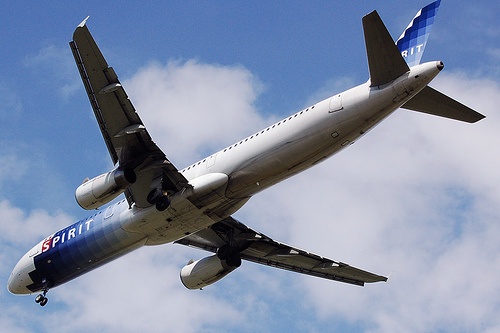}
   &
   \includegraphics[width=0.2\textwidth]{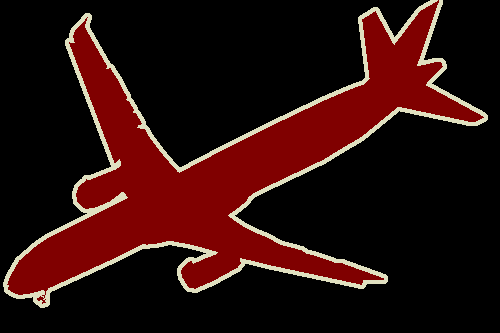}
   &
  \includegraphics[width=0.2\textwidth]{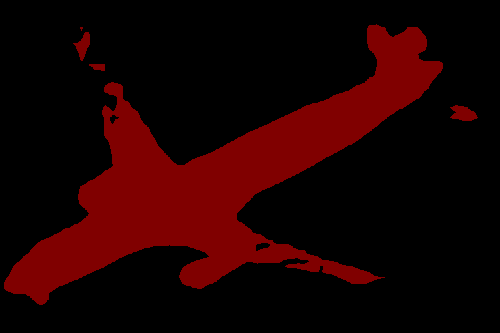}
  &
  {\includegraphics[width=0.2\textwidth]{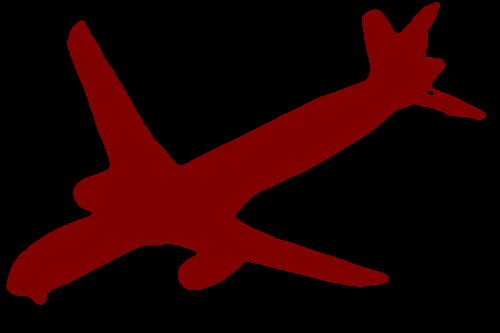}} \\
   \includegraphics[width=0.2\textwidth]{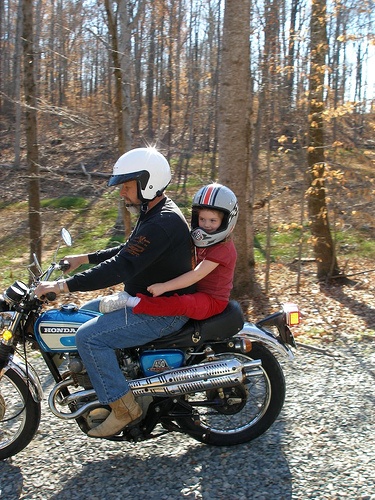}
   &
   \includegraphics[width=0.2\textwidth]{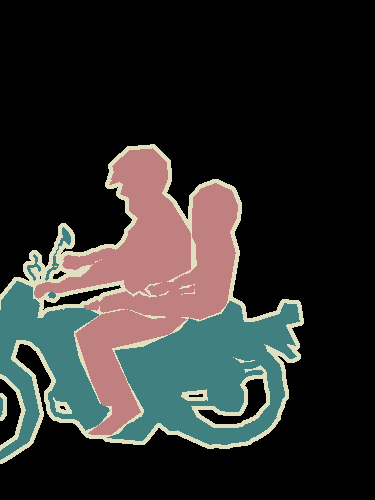}
   &
  \includegraphics[width=0.2\textwidth]{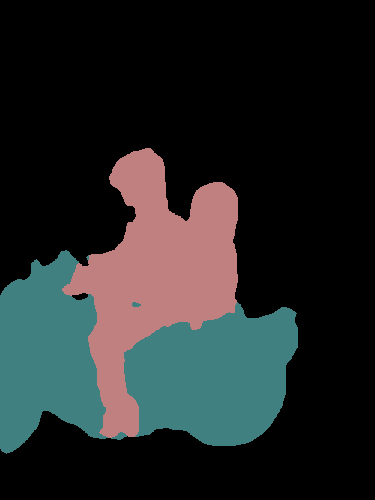}
  &
  {\includegraphics[width=0.2\textwidth]{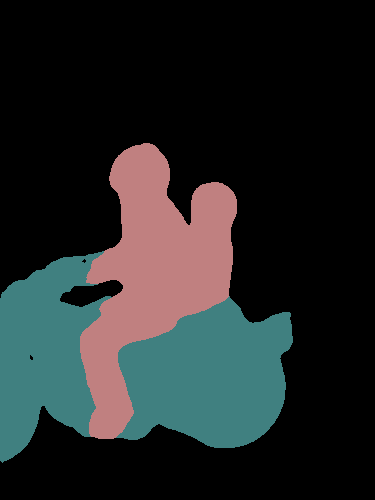}} \\
                  image & ground-truth & FCN-8s \cite{long15_cvpr} & LRN \Tstrut\\
 \end{tabular}}
    \caption{Qualitative comparisons between the FCN\cite{long15_cvpr} model and our LRN model on the PASCAL VOC 2012 dataset.}
      \label{fig:val_pascal}
    \end{center}
 \end{figure*}
\subsection{Implementation Details and Baselines}\label{sec:implementation}
Following \cite{badrinarayanan15_arxiv}, we train our network on the CamVid and SUN RGB-D datasets using the class balancing technique in \cite{eigen15_iccv} as there is a large variation in the number of pixels for each category. For example, most of the pixels on the CamVid dataset are labeled as roads and cars. Each class is assigned a weight in the loss function. The weights are calculated by taking the ratio of the median of the class frequencies to the individual class frequencies from the training dataset. For the CamVid and SUN RGB-D datasets, training images are resized to a common size of $360\times 480$ pixels. For the PASCAL VOC 2012 dataset, when fed into LRN, we randomly crop a $320\times 320$ region from each image during training and subtracted a mean pixel provided by VGG net at each position. During test, we generate a $512 \times 512$ segmentation map and crop the final label out from it similar to \cite{chen15_iclr}. 

We implemented LRN using Caffe \cite{jia14_caffe}. A GTX Titan X GPU was used both in training and testing for acceleration. The whole network was trained end-to-end by using back propagation algorithm with the following hyper-parameters: mini-batch size (10), learning rate (0.001), momentum (0.9), weight decay (0.0005), number of iterations ($\sim80,000$). After each $50,000$ iterations we reduce the learning rate by a factor of $0.1$. The encoder network is fine-tuned from a pre-trained VGG-16 network.

To demonstrate the merit of individual component of our model, we perform an ablation study by comparing with the following baselines.

{\bf SegNet}: This is the SegNet model in \cite{badrinarayanan15_arxiv}. Whenever possible, we use the reported numbers in \cite{badrinarayanan15_arxiv} in our comparison.

{\bf SegNet+Deep Supervision~(SegNet+DS)}: This is the SegNet model with multiple loss functions to provide deep supervisions.
 \begin{figure*}[ht!]
    \begin{center}
    \setlength\tabcolsep{0.7pt}
    \def\arraystretch{0.5}
    \resizebox{\textwidth}{!}{
 \begin{tabular}{*{8}{c }}
    \includegraphics[width=0.2\textwidth]{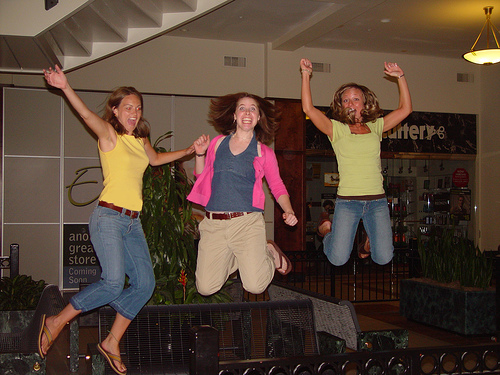}
   &
   \includegraphics[width=0.2\textwidth]{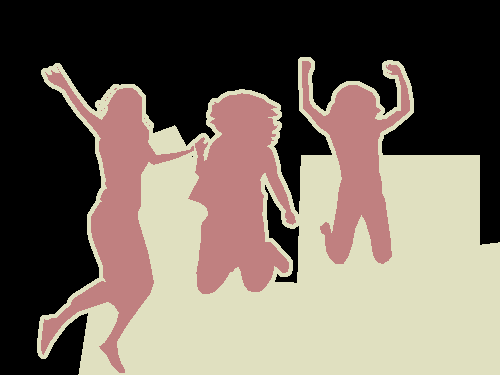}
   &
  \includegraphics[width=0.2\textwidth]{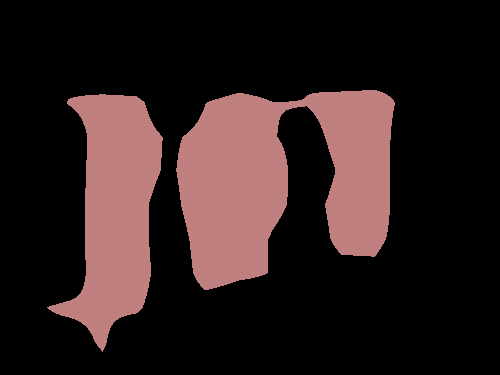}
  &
  \includegraphics[width=0.2\textwidth]{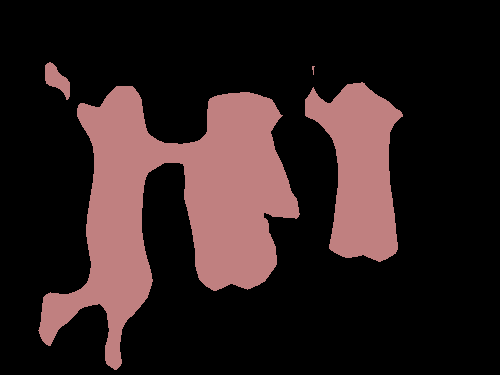}
  &
    \includegraphics[width=0.2\textwidth]{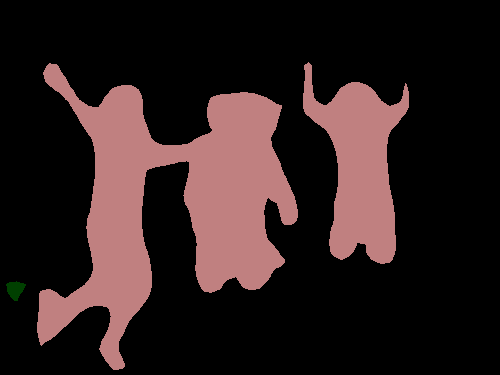}
  &
   \includegraphics[width=0.2\textwidth]{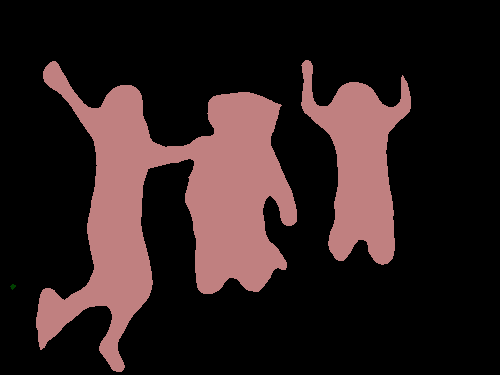}
  &
   \includegraphics[width=0.2\textwidth]{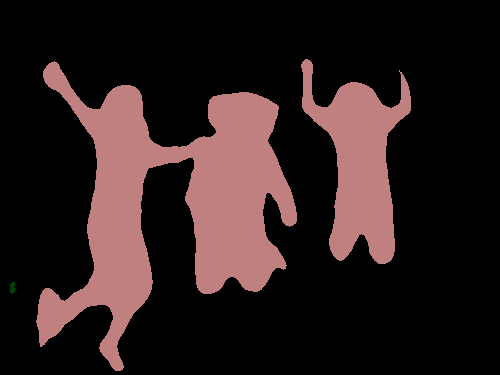}
  &
 \includegraphics[width=0.2\textwidth]{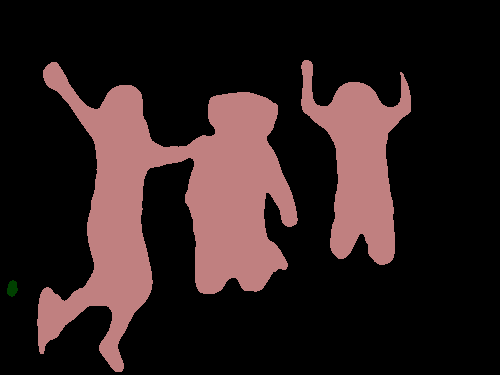} \\
   \includegraphics[width=0.2\textwidth]{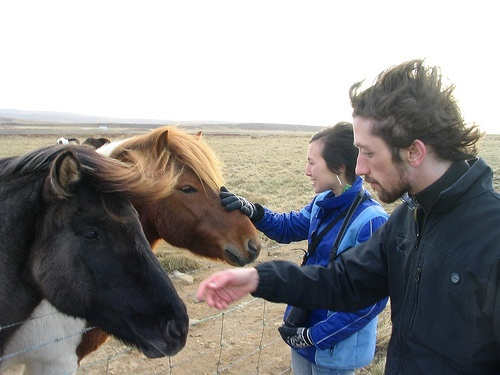}
   &
   \includegraphics[width=0.2\textwidth]{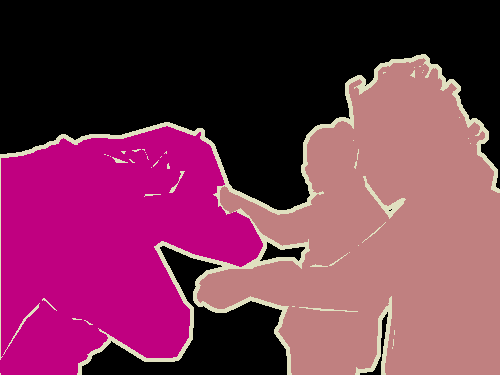}
   &
  \includegraphics[width=0.2\textwidth]{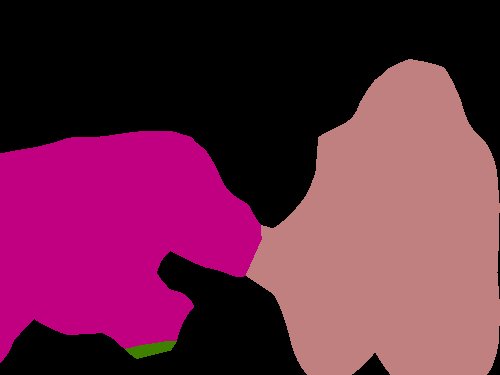}
  &
  \includegraphics[width=0.2\textwidth]{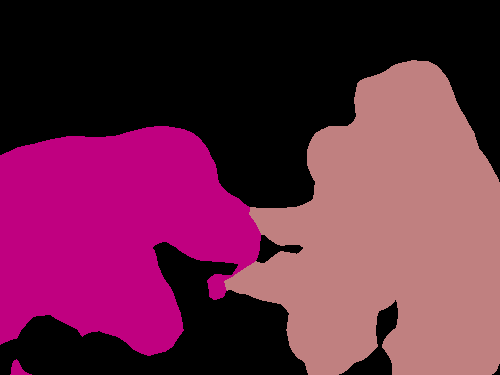}
  &
    \includegraphics[width=0.2\textwidth]{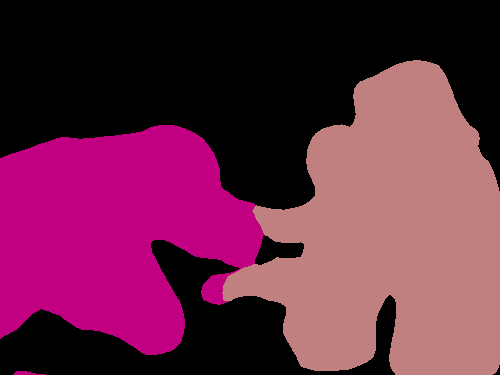}
  &
   \includegraphics[width=0.2\textwidth]{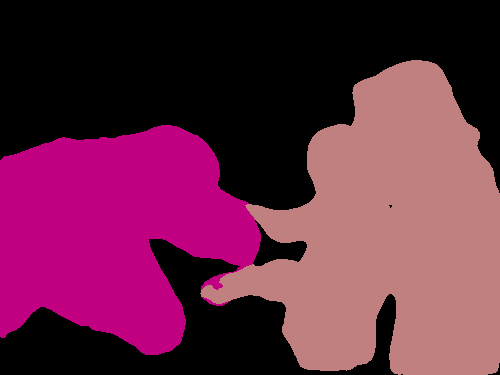}
  &
   \includegraphics[width=0.2\textwidth]{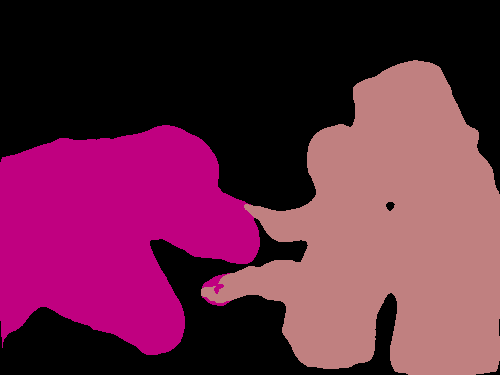}
  &
 \includegraphics[width=0.2\textwidth]{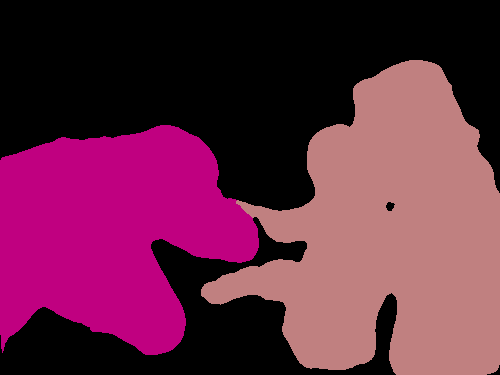}\\
   \includegraphics[width=0.2\textwidth]{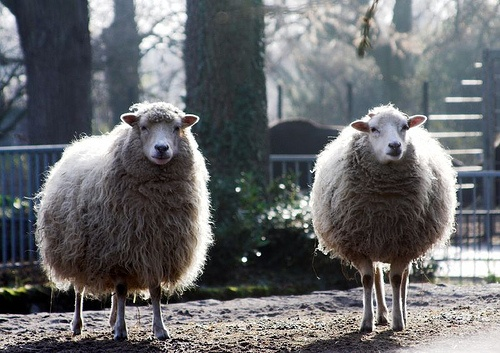}
   &
   \includegraphics[width=0.2\textwidth]{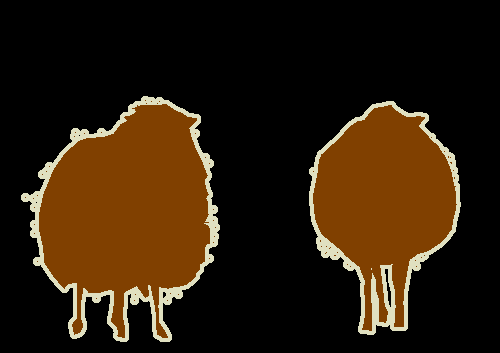}
   &
  \includegraphics[width=0.2\textwidth]{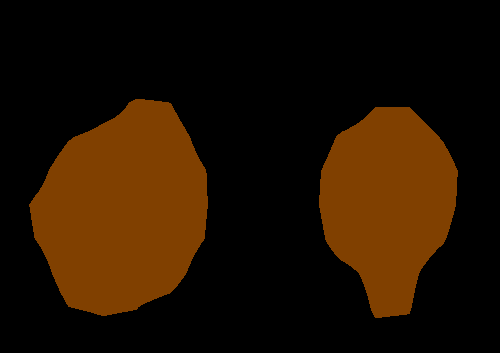}
  &
  \includegraphics[width=0.2\textwidth]{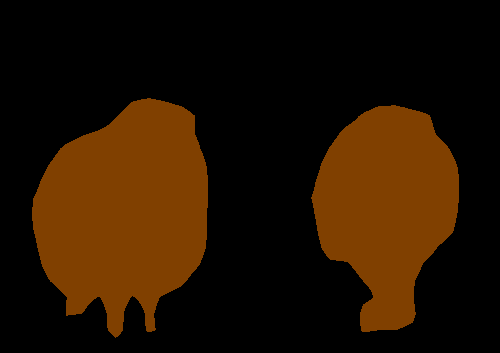}
  &
    \includegraphics[width=0.2\textwidth]{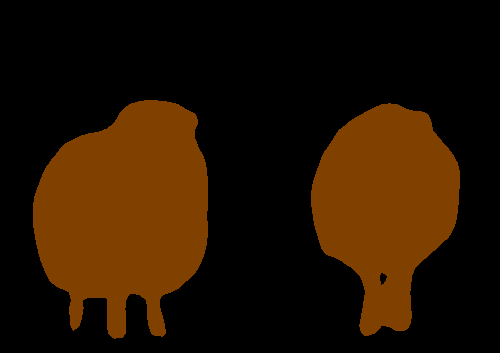}
  &
   \includegraphics[width=0.2\textwidth]{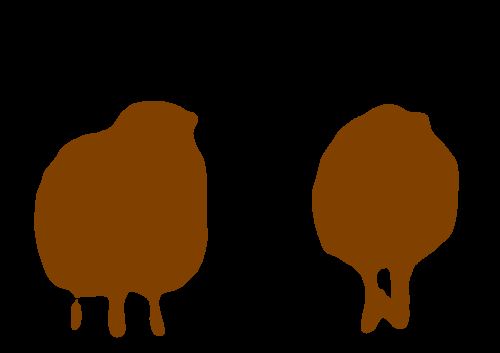}
  &
   \includegraphics[width=0.2\textwidth]{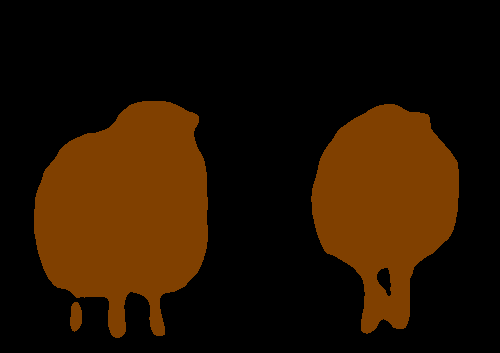}
  &
   \includegraphics[width=0.2\textwidth]{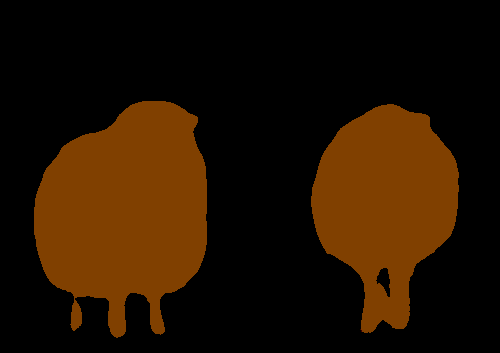}\\
   \includegraphics[width=0.2\textwidth]{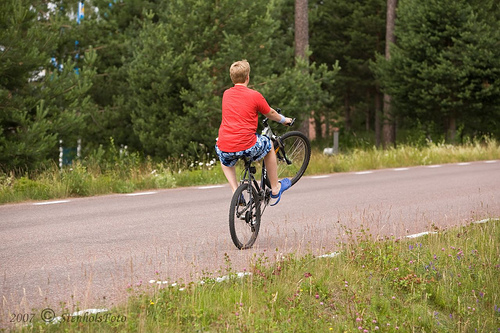}
   &
   \includegraphics[width=0.2\textwidth]{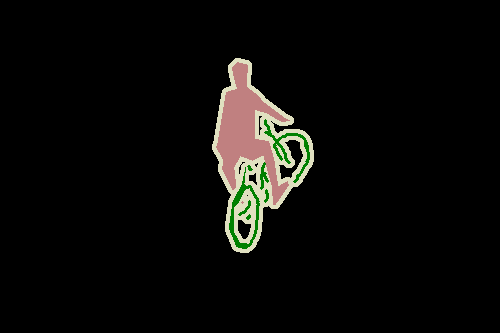}
   &
  \includegraphics[width=0.2\textwidth]{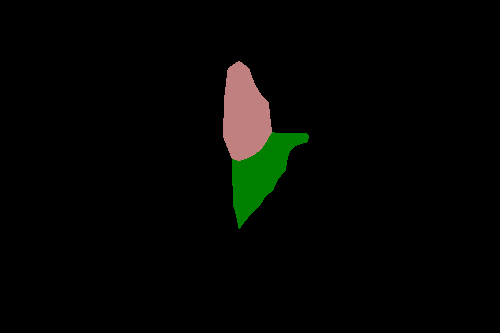}
  &
  \includegraphics[width=0.2\textwidth]{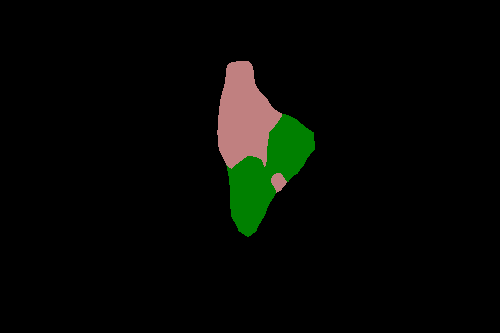}
  &
    \includegraphics[width=0.2\textwidth]{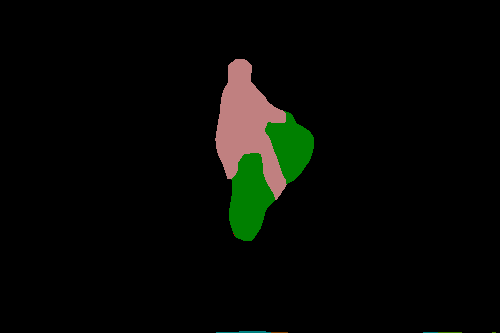}
  &
   \includegraphics[width=0.2\textwidth]{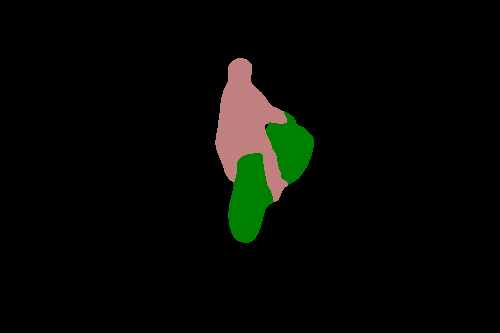}
  &
   \includegraphics[width=0.2\textwidth]{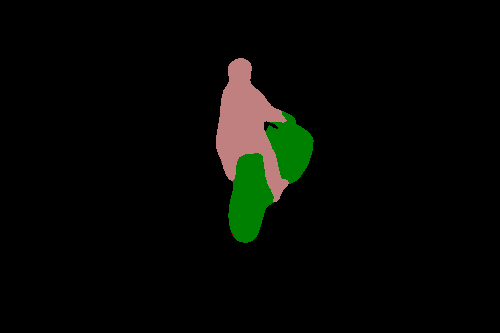}
  &
   \includegraphics[width=0.2\textwidth]{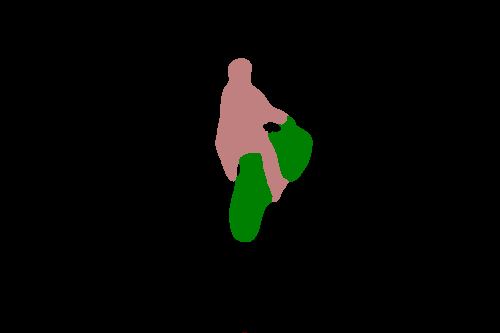}\\
   image & ground-truth & $s_1$ & $s_2$ & $s_3$ & $s_4$ & $s_5$ &$s_6$\Tstrut\\
 \end{tabular}}
    \caption{Stage-wise visualization of semantic segmentation results on PASCAL VOC 2012. We show the segmentation result obtained from each of the 6 label maps $s_k$ ($k=1,2,...,6$) produced by the decoder of our model.}
    \label{fig:stage_pascal}
 \end{center}
 \end{figure*}
\begin{table*}[ht!]
\resizebox{\textwidth}{!}{%
    \centering
    \begin{tabular}{l|c|c|c|c|c|c|c|c|c|c|c|c|c}
        Method&\multicolumn{1}{c}{\rot{Building}}&\multicolumn{1}{c}{\rot{Tree}}&\multicolumn{1}{c}{\rot{Sky}}&\multicolumn{1}{c}{\rot{Car}}&\multicolumn{1}{c}{\rot{Sign}}&\multicolumn{1}{c}{\rot{Road}}&\multicolumn{1}{c}{\rot{Pedestrian}}&\multicolumn{1}{c}{\rot{Fence}}&\multicolumn{1}{c}{\rot{Pole}}&\multicolumn{1}{c}{\rot{Sidewalk}}&\multicolumn{1}{c|}{\rot{Bicyclist}} & \rot{Class avg} & \rot{Mean IoU}\\
       \hline
       \hline
       \cite{zhang10_eccv}&85.3&57.3&95.4&69.2&46.5&{\bf 98.5}&23.8&44.3&22.0&38.1&28.7&55.4&-\\
       \hline
       \cite{tighe13_ijcv}&87.0&67.1&{\bf 96.9}&62.7&30.1&95.9&14.7&17.9&1.7&70.0&19.4&51.2&-\\
       \hline
       \cite{ladicky10_eccv}&81.5&76.6&96.2&78.7&40.2&93.9&43.0&47.6&14.3&81.5&33.9&62.5&-\\
       \hline
       \hline
       SegNet~\cite{badrinarayanan15_arxiv}&88.0&87.3&92.3&80.0&29.5&97.6&57.2&49.4&27.8&84.8&30.7&65.9&50.2\\
       SegNet+DS&87.0&85.1&77.1&79.8& 56.3&94.8&71.1&50.5&39.7&88.6&50.6& 70.1&53.7 \\
        \textbf{LRN}&{\bf 89.8}&{\bf 88.1}&78.5&{\bf 86.3}&{\bf 61.2}&96.8&{\bf 82.1}&{\bf 59.0}&{\bf 45.4}&{\bf 92.6}&{\bf 69.7}&\textbf{77.2}&\textbf{61.7}  \\
       \hline
    \end{tabular}}
\caption{Quantitative results on the CamVid~\cite{brostow09_prl} dataset. For each method, we report the accuracy for each class and the average class accuracy. We also report the mean IoU of all classes if it is available. Our model (LRN) outperforms all the other methods. }
  \label{table:quant_camvid}
\end{table*}
\subsection{PASCAL VOC 2012}\label{sec:pascal}
PASCAL VOC 2012 is a challenging segmentation benchmark~\cite{everingham10_ijcv}. It consists of 20 object categories and the background. Following previous works such as ~\cite{chen15_iclr, long15_cvpr, badrinarayanan15_arxiv}, we employ 10,582 images for training, 1,449 images for validation, and 1,456 images for testing. We train our model on the trainval set and evaluate our model on the test set. Quantitative results are shown in Table \ref{tab:quant_pascal}. We can see that SegNet+DS performs better than SegNet. This shows that deep supervisions (i.e. multiple losses) help boosting the performance. In addition, Table \ref{tab:quant_pascal} shows that our model (LRN) outperforms SegNet+DS. This demonstrates that the refinement modules give additional performance improvement. Note that we also perform better than the popular FCN~\cite{long15_cvpr} model for semantic segmentation. We present qualitative results in Fig.~\ref{fig:val_pascal}. The results show that LRN can preserve subtle details of objects. To further illustrate the effectiveness of the coarse-to-fine segmentation in our model, Fig.~\ref{fig:stage_pascal} shows each of the label maps produced by the refinement network in our model. We can see that the coarse-to-fine refinement scheme can progressively improve the details of segmentation maps by recovering the missing parts (e.g., the leg of the person and sheep in the top and $3^{rd}$ row respectively). 

\subsection{CamVid}\label{sec:camvid}
\begin{figure*}
    \begin{center}
    \setlength\tabcolsep{1pt}
      \def\arraystretch{0.5}
    \resizebox{\textwidth}{!}{
 \begin{tabular}{*{4}{c }}
    \includegraphics[width=0.2\textwidth]{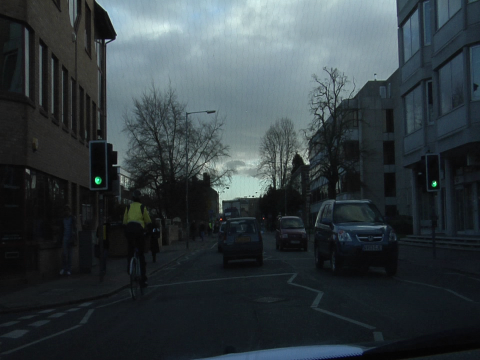}
   &
   \includegraphics[width=0.2\textwidth]{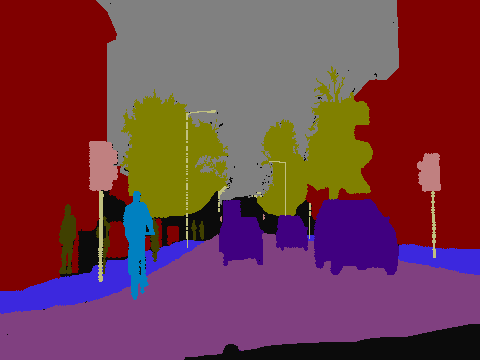}
   &
  \includegraphics[width=0.2\textwidth]{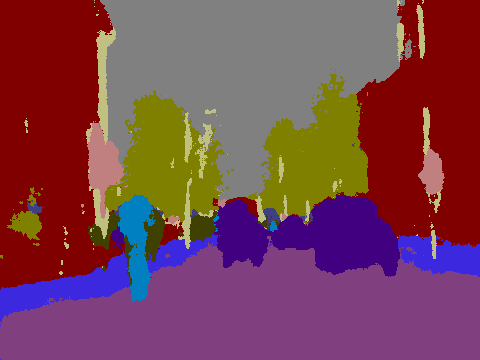}
  &
  {\includegraphics[width=0.2\textwidth]{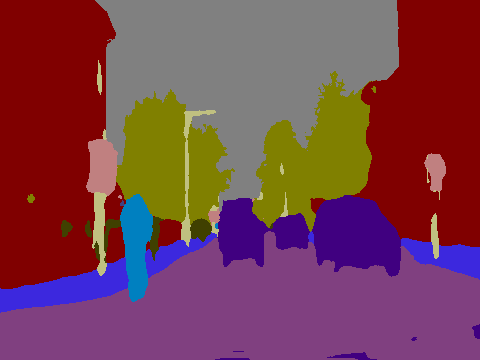}} \\
  \includegraphics[width=0.2\textwidth]{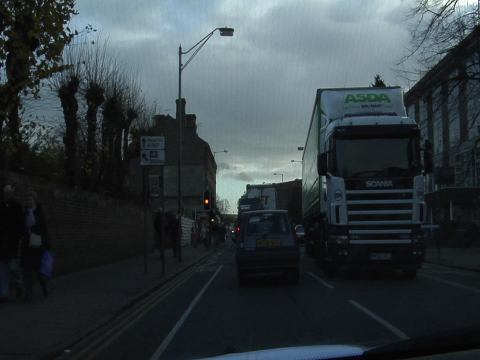}
   &
   \includegraphics[width=0.2\textwidth]{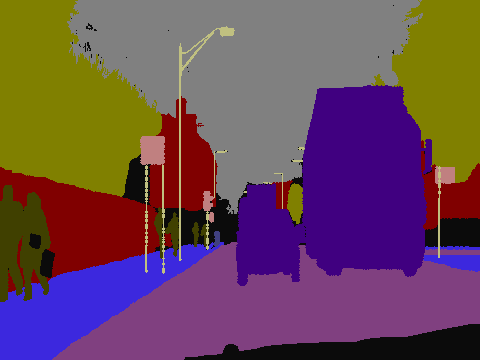}
   &
  \includegraphics[width=0.2\textwidth]{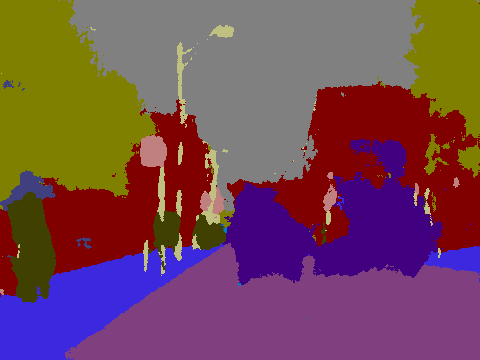}
  &
  {\includegraphics[width=0.2\textwidth]{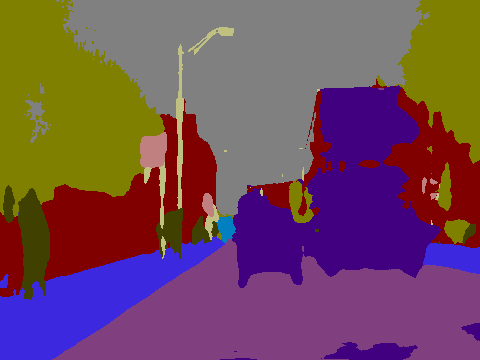}}\\
  \includegraphics[width=0.2\textwidth]{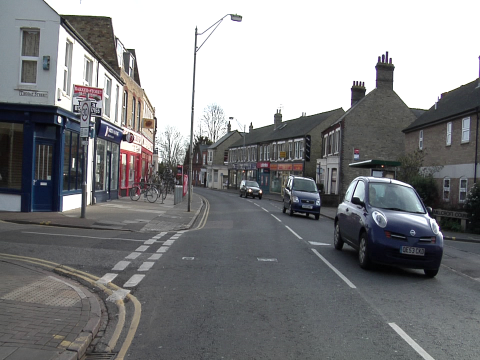}
   &
   \includegraphics[width=0.2\textwidth]{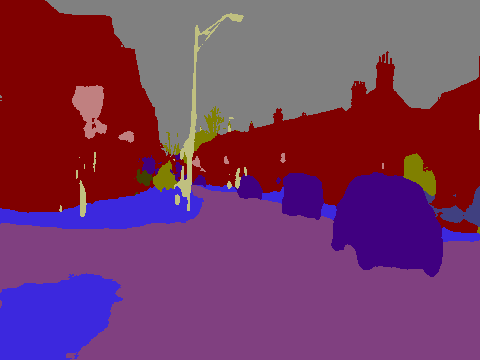}
   &
  \includegraphics[width=0.2\textwidth]{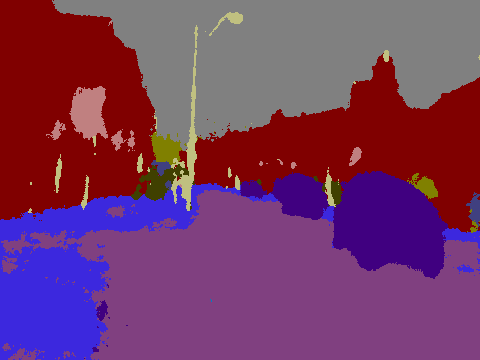}
  &
  {\includegraphics[width=0.2\textwidth]{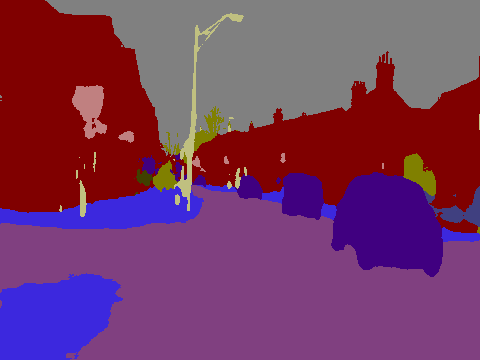}}\\
  \includegraphics[width=0.2\textwidth]{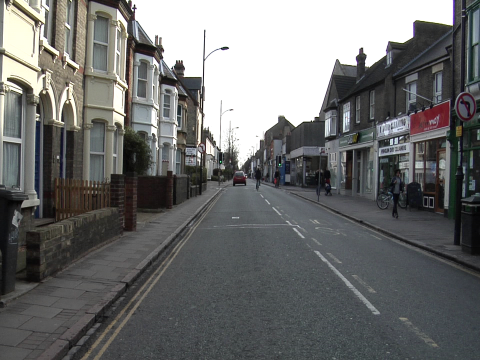}
   &
   \includegraphics[width=0.2\textwidth]{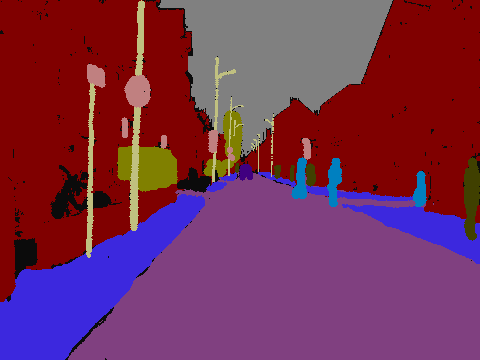}
   &
  \includegraphics[width=0.2\textwidth]{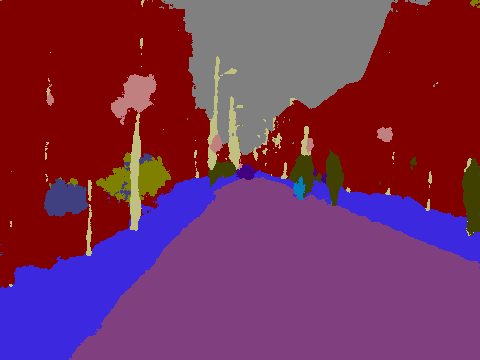}
  &
  {\includegraphics[width=0.2\textwidth]{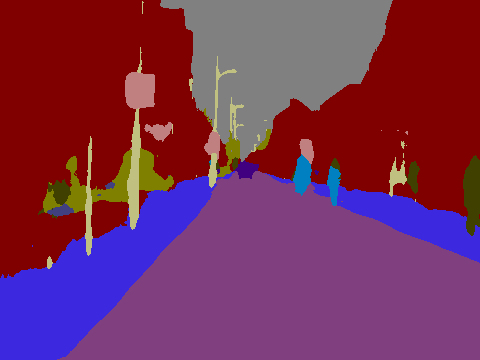}}\\
                  image & ground-truth & SegNet \cite{badrinarayanan15_arxiv} & LRN \Tstrut\\
  \end{tabular}
  \begin{tabular}{*{4}{c }}
   \includegraphics[width=0.2\textwidth]{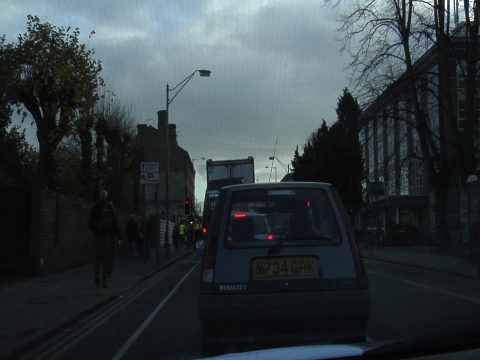}
   &
   \includegraphics[width=0.2\textwidth]{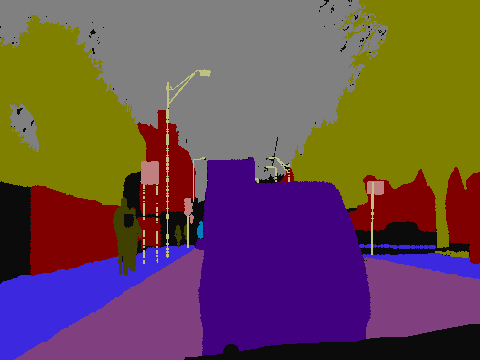}
   &
  \includegraphics[width=0.2\textwidth]{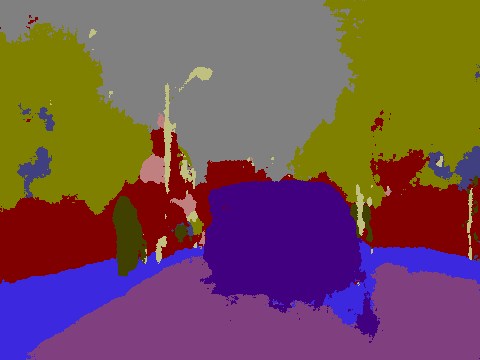}
  &
  {\includegraphics[width=0.2\textwidth]{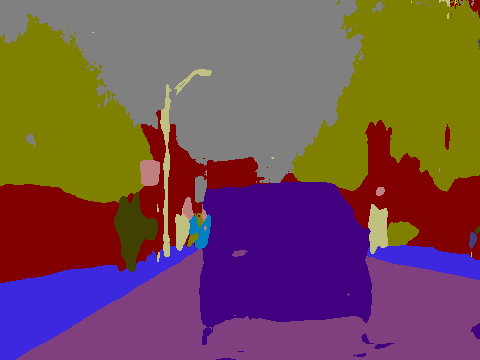}}\\
  \includegraphics[width=0.2\textwidth]{images/camvid/original/0_92.png}
   &
   \includegraphics[width=0.2\textwidth]{images/camvid/gt/92_segnet.png}
   &
  \includegraphics[width=0.2\textwidth]{images/camvid/segnet/92_segnet.png}
  &
  {\includegraphics[width=0.2\textwidth]{images/camvid/skip/92_segnet.png}}\\
  \includegraphics[width=0.2\textwidth]{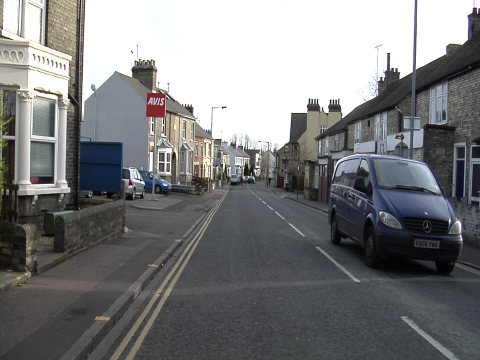}
   &
   \includegraphics[width=0.2\textwidth]{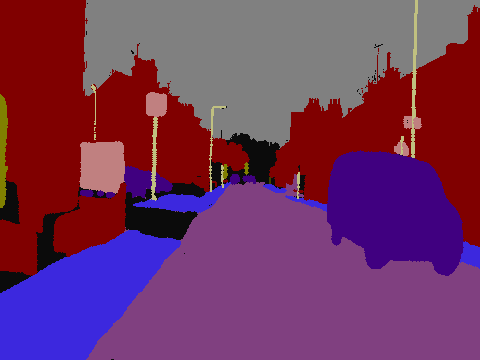}
   &
  \includegraphics[width=0.2\textwidth]{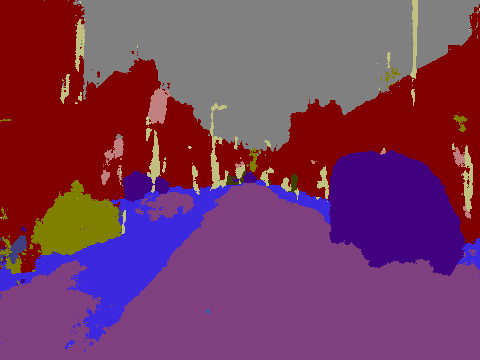}
  &
  {\includegraphics[width=0.2\textwidth]{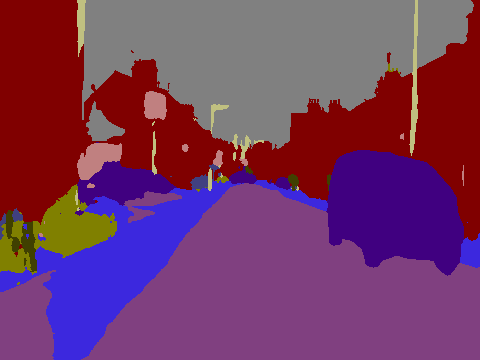}}\\
   \includegraphics[width=0.2\textwidth]{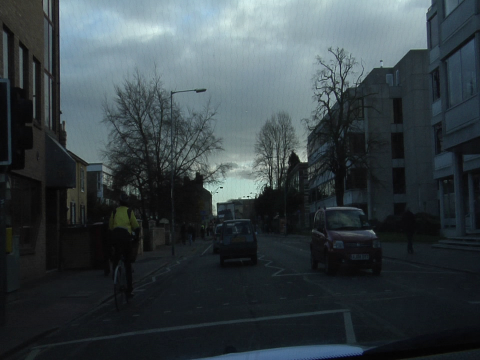}
   &
   \includegraphics[width=0.2\textwidth]{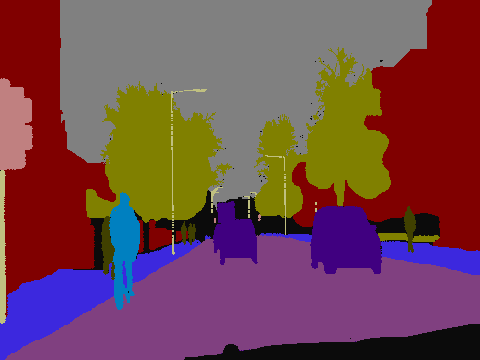}
   &
  \includegraphics[width=0.2\textwidth]{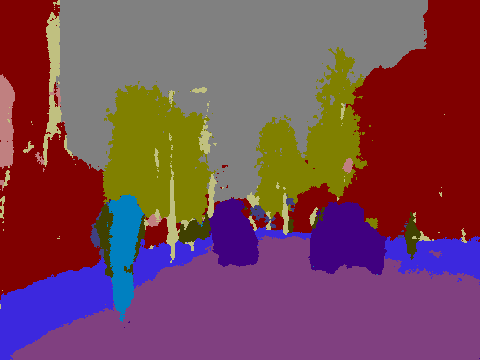}
  &
  {\includegraphics[width=0.2\textwidth]{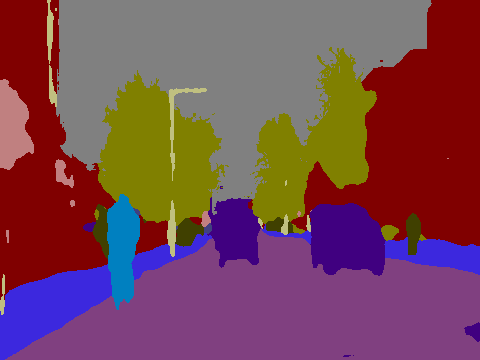}}\\
                  image & ground-truth & SegNet \cite{badrinarayanan15_arxiv} & LRN \Tstrut\\
 \end{tabular}}
    \caption{Qualitative results on the CamVid dataset. LRN predictions capable of retaining the shape of smaller object categories (e.g. column-pole, side-walk, bicyclist, and sign-symbols) accurately compared to SegNet \cite{badrinarayanan15_arxiv}. } 
    \label{fig:eval_camvid}
    \end{center}
 \end{figure*}

The CamVid dataset~\cite{brostow09_prl} consists of a ten minutes video footage with 701 images. The video footage was recorded from driving around various urban settings and is particularly challenging given variable lighting settings (e.g. dusk, dawn). The dataset provides ground-truth labels that associate each pixel with one of 32 semantic class. Following \cite{badrinarayanan15_arxiv}, we consider 11 larger semantic classes (road, building, sky, tree, sidewalk, car, column-pole, fence, pedestrian, bicyclist, and sign-symbol) among the 32 semantic classes in our experiment. Table~\ref{table:quant_camvid} shows the comparison of our model with other state-of-the-art approaches on this dataset. For each method, we report the accuracy for each class and the average class accuracy. We also report the mean IoU score if it is available. Again, our LRN model outperforms SegNet+DS, which in turn outperforms SegNet. We show some qualitative results on this dataset in Fig.~\ref{fig:eval_camvid}.
 \subsection{SUN RGB-D}\label{sec:sun}
\begin{figure*}[ht!]
    \begin{center}
    \setlength\tabcolsep{1pt}
      \def\arraystretch{0.5}
    \resizebox{\textwidth}{!}{
  \begin{tabular}{cccc}
   \includegraphics[width=0.2\textwidth]{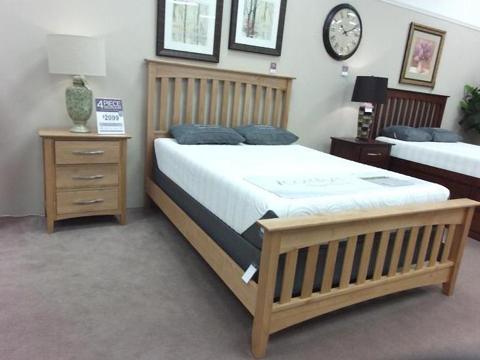}
   &
   \includegraphics[width=0.2\textwidth]{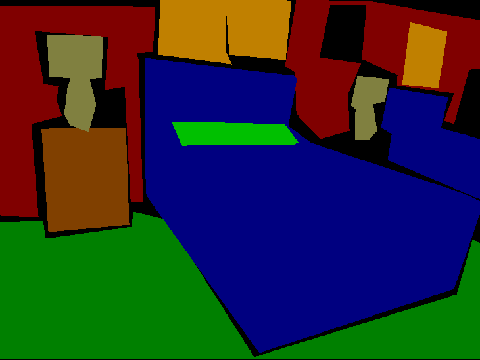}
   &
  \includegraphics[width=0.2\textwidth]{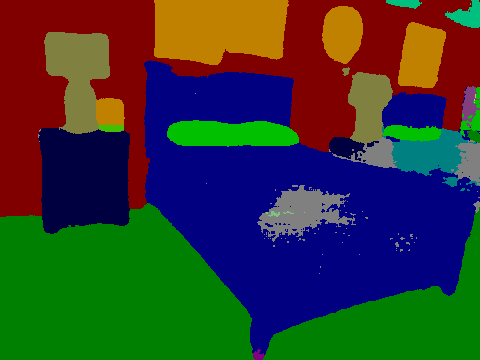}
  &
  {\includegraphics[width=0.2\textwidth]{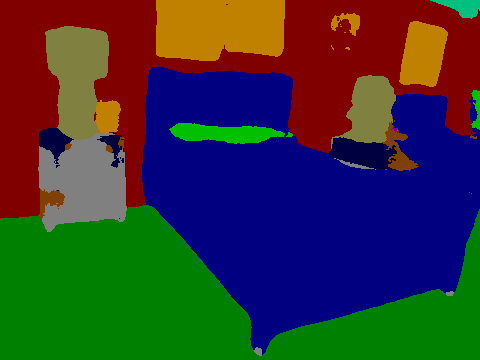}} \\
  \includegraphics[width=0.2\textwidth]{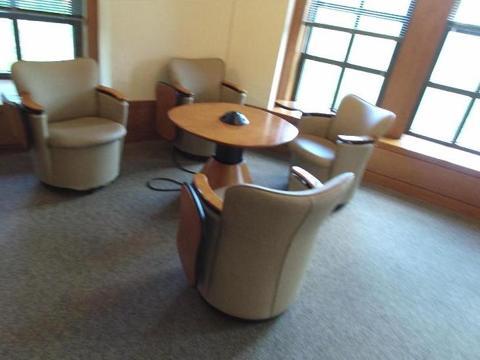}
   &
   \includegraphics[width=0.2\textwidth]{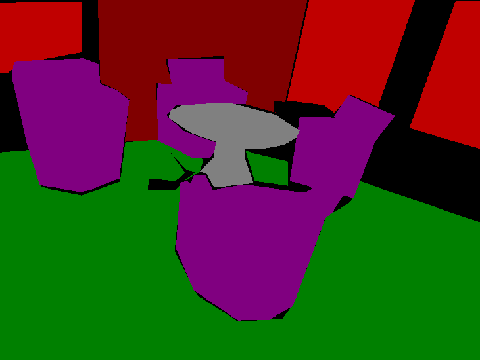}
   &
  \includegraphics[width=0.2\textwidth]{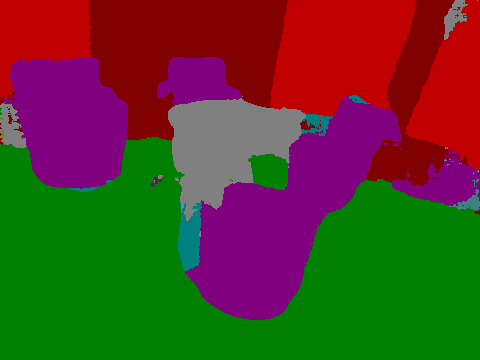}
  &
  {\includegraphics[width=0.2\textwidth]{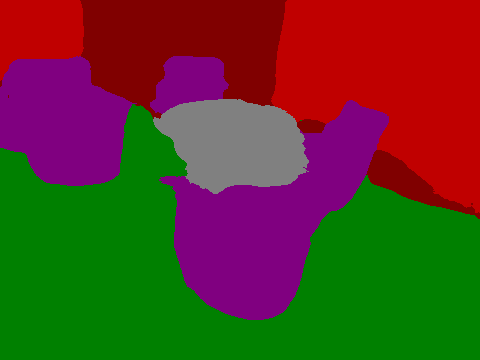}} \\
  \includegraphics[width=0.2\textwidth]{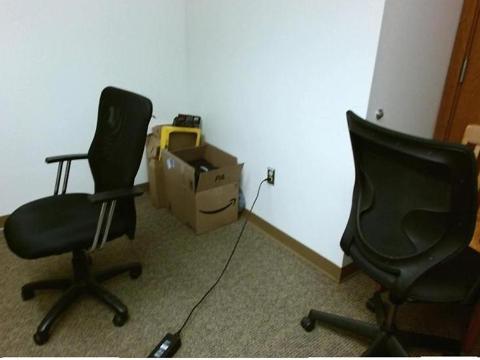}
   &
   \includegraphics[width=0.2\textwidth]{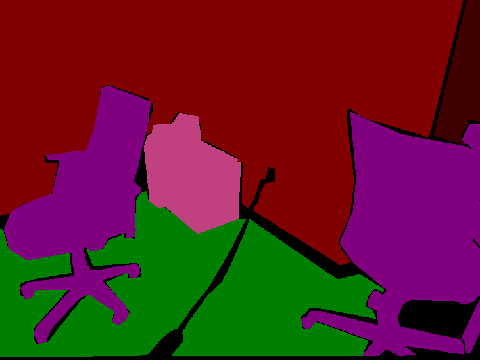}
   &
  \includegraphics[width=0.2\textwidth]{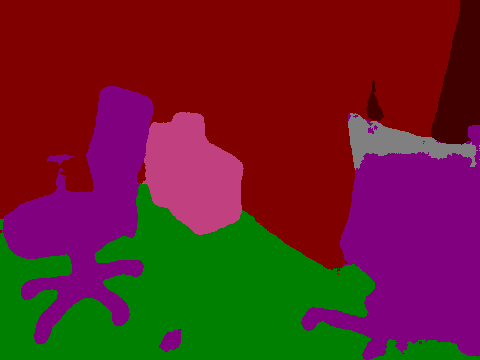}
  &
  {\includegraphics[width=0.2\textwidth]{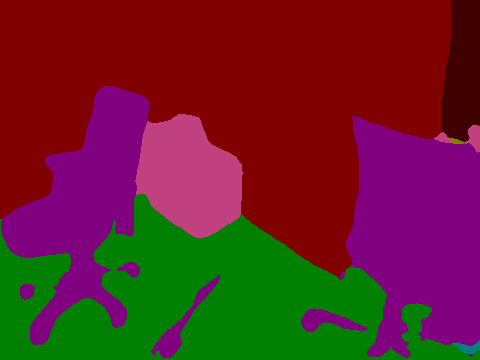}} \\
   image & ground-truth & SegNet \cite{badrinarayanan15_arxiv} & LRN \Tstrut\\
   \end{tabular}
   \begin{tabular}{cccc}
   \includegraphics[width=0.2\textwidth]{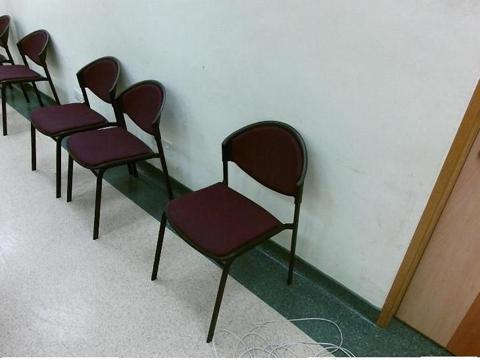}
   &
   \includegraphics[width=0.2\textwidth]{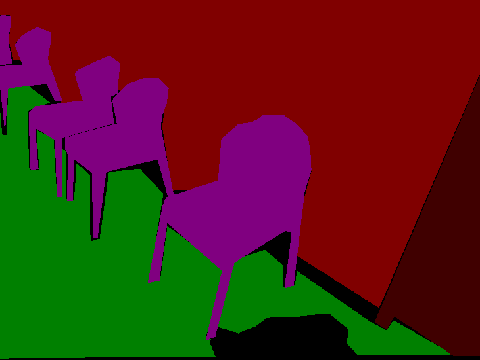}
   &
  \includegraphics[width=0.2\textwidth]{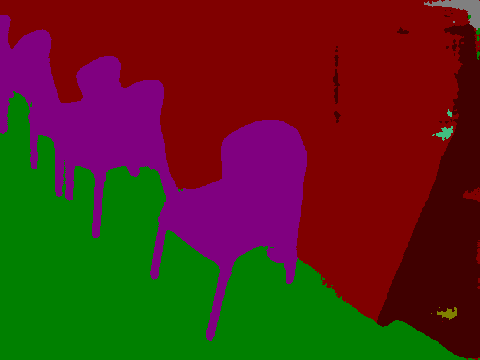}
  &
  {\includegraphics[width=0.2\textwidth]{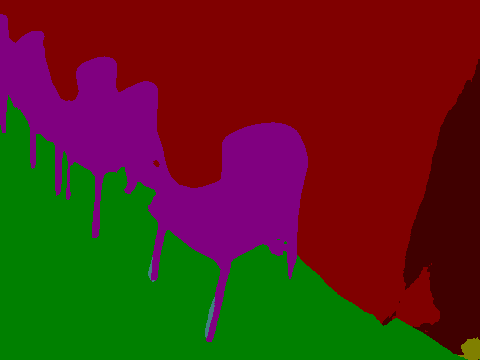}} \\
  \includegraphics[width=0.2\textwidth]{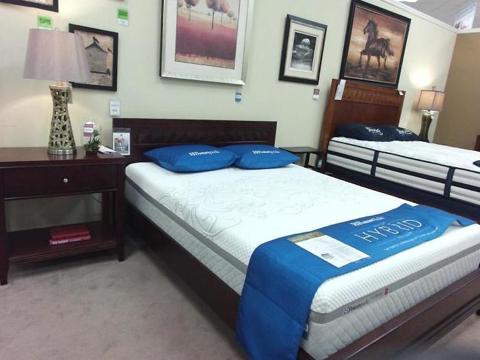}
   &
   \includegraphics[width=0.2\textwidth]{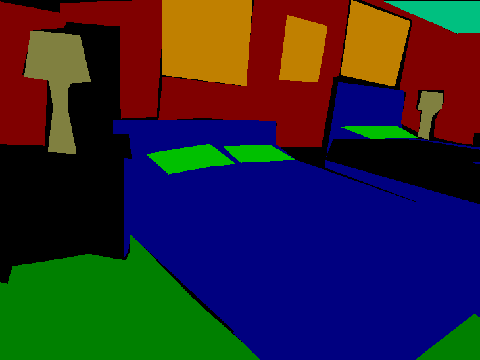}
   &
  \includegraphics[width=0.2\textwidth]{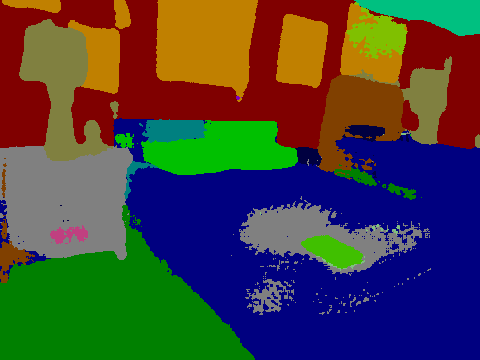}
  &
  {\includegraphics[width=0.2\textwidth]{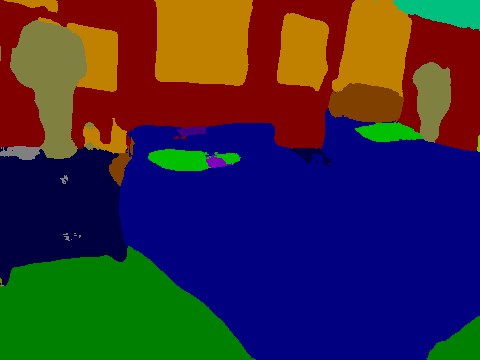}} \\
  \includegraphics[width=0.2\textwidth]{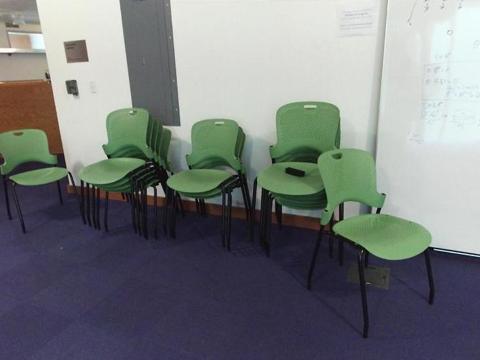}
   &
   \includegraphics[width=0.2\textwidth]{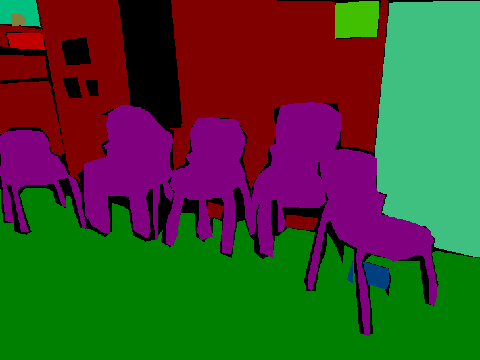}
   &
  \includegraphics[width=0.2\textwidth]{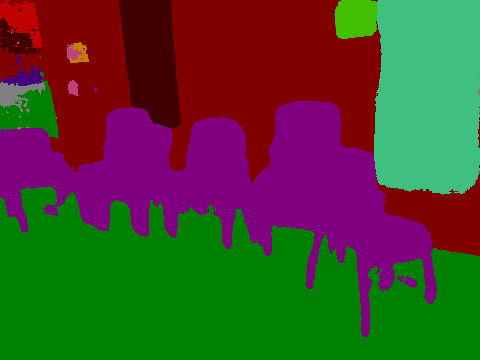}
  &
  {\includegraphics[width=0.2\textwidth]{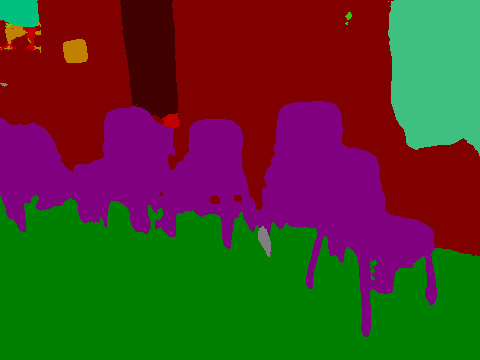}} \\
   image & ground-truth & SegNet \cite{badrinarayanan15_arxiv} & LRN \Tstrut\\
   \end{tabular}}
    \caption{Qualitative results on the SUN RGB-D dataset.} 
    \label{fig:sun_example}
    \end{center}
 \end{figure*}

 We also evaluate our model on the SUN RGB-D dataset~\cite{song15_cvpr} which contains 5,285 training and 5,050 test images. The images consist of indoor scenes of varying resolution and aspect ratio. There are 37 indoor scene classes with corresponding segmentation labels (background is not considered as a class and is ignored during training and testing). The segmentation labels are instance-wise, i.e. multiple instances of same class in an image have different labels. We convert the ground-truth labels into class-wise segmentation labels so that all instances of the same class have the same corresponding label. Although the dataset also contains depth information, we only use the RGB images to train and test our model. Quantitative results on this dataset are shown in Table~\ref{tab:quant_cat_sun}. Our LRN model achieves better mean IoU than other baselines on this dataset. We show some qualitative results in Fig.~\ref{fig:sun_example}.

 \begin{table*}[ht!]
\resizebox{\textwidth}{!}{%
    \centering
  \def\arraystretch{2.5}
    \begin{tabular}{r|ccccccccccccccccccccccccccccccccccccc|c|c}
     Method&\multicolumn{1}{c}{\rot{Wall}}&\multicolumn{1}{c}{\rot{Floor}}&\multicolumn{1}{c}{\rot{Cabinet}}&\multicolumn{1}{c}{\rot{Bed}}&\multicolumn{1}{c}{\rot{Chair}}&\multicolumn{1}{c}{\rot{Sofa}}&\multicolumn{1}{c}{\rot{Table}}&\multicolumn{1}{c}{\rot{Door}}&\multicolumn{1}{c}{\rot{Window}}&\multicolumn{1}{c}{\rot{Bookshelf}}&\multicolumn{1}{c}{\rot{Picture}}&\multicolumn{1}{c}{\rot{Counter}}&\multicolumn{1}{c}{\rot{Blinds}}&\multicolumn{1}{c}{\rot{Desk}}&\multicolumn{1}{c}{\rot{Shelves}}&\multicolumn{1}{c}{\rot{Curtain}}&\multicolumn{1}{c}{\rot{Dresser}}&\multicolumn{1}{c}{\rot{Pillow}}&\multicolumn{1}{c}{\rot{Mirror}}&\multicolumn{1}{c}{\rot{Floor Mat}} &\multicolumn{1}{c}{\rot{Clothes}}&\multicolumn{1}{c}{\rot{Ceiling}}&\multicolumn{1}{c}{\rot{Books}}&\multicolumn{1}{c}{\rot{Fridge}}&\multicolumn{1}{c}{\rot{TV}}&\multicolumn{1}{c}{\rot{Paper}}&\multicolumn{1}{c}{\rot{Towel}}&\multicolumn{1}{c}{\rot{Curtain}}&\multicolumn{1}{c}{\rot{Box}}&\multicolumn{1}{c}{\rot{Whiteboard}}&\multicolumn{1}{c}{\rot{Person}}&\multicolumn{1}{c}{\rot{Night Stand}}&\multicolumn{1}{c}{\rot{Toilet}}&\multicolumn{1}{c}{\rot{Sink}}&\multicolumn{1}{c}{\rot{Lamp}}&\multicolumn{1}{c}{\rot{Bathtub}}&\multicolumn{1}{c|}{\rot{Bag}}& \rot{Class avg}&\rot{mean IoU}\\
       \hline
  \cite{kendall15_arxiv} &\multicolumn{1}{c}{\rot{80.2}}&\multicolumn{1}{c}{\rot{90.9}}&\multicolumn{1}{c}{\rot{{\bf 58.9}}} &\multicolumn{1}{c}{\rot{64.8}} &\multicolumn{1}{c}{\rot{76.0}}&\multicolumn{1}{c}{\rot{58.6}}&\multicolumn{1}{c}{\rot{62.6}}&\multicolumn{1}{c}{\rot{47.7}}&\multicolumn{1}{c}{\rot{{\bf 66.4}}}&\multicolumn{1}{c}{\rot{31.2}}&\multicolumn{1}{c}{\rot{63.6}}&\multicolumn{1}{c}{\rot{33.8}}&\multicolumn{1}{c}{\rot{{\bf 46.7}}}&\multicolumn{1}{c}{\rot{19.7}}&\multicolumn{1}{c}{\rot{16.2}}&\multicolumn{1}{c}{\rot{{\bf 67.0}}}&\multicolumn{1}{c}{\rot{42.3}}&\multicolumn{1}{c}{\rot{{\bf 57.1}}}&\multicolumn{1}{c}{\rot{39.1}}&\multicolumn{1}{c}{\rot{0.1}}&\multicolumn{1}{c}{\rot{24.4}}&\multicolumn{1}{c}{\rot{{\bf 84.0}}}&\multicolumn{1}{c}{\rot{48.7}}&\multicolumn{1}{c}{\rot{21.3}}&\multicolumn{1}{c}{\rot{49.5}}&\multicolumn{1}{c}{\rot{30.6}}&\multicolumn{1}{c}{\rot{18.8}}&\multicolumn{1}{c}{\rot{0.1}}&\multicolumn{1}{c}{\rot{24.1}}&\multicolumn{1}{c}{\rot{56.8}}&\multicolumn{1}{c}{\rot{17.9}}&\multicolumn{1}{c}{\rot{42.9}}&\multicolumn{1}{c}{\rot{73.0}}&\multicolumn{1}{c}{\rot{66.2}}&\multicolumn{1}{c}{\rot{48.8}}&\multicolumn{1}{c}{\rot{45.1}}&\multicolumn{1}{c|}{\rot{24.1}}&{\rot{45.9}}&\multicolumn{1}{c}{\rot{30.7}}\\
       \hline

      SegNet~\cite{badrinarayanan15_arxiv}&\multicolumn{1}{c}{\rot{{\bf 86.6|}}}&\multicolumn{1}{c}{\rot{92.0}}&\multicolumn{1}{c}{\rot{52.4}} &\multicolumn{1}{c}{\rot{{\bf 68.4}}} &\multicolumn{1}{c}{\rot{76.0}}&\multicolumn{1}{c}{\rot{54.3}}&\multicolumn{1}{c}{\rot{59.3}}&\multicolumn{1}{c}{\rot{37.4}}&\multicolumn{1}{c}{\rot{53.8}}&\multicolumn{1}{c}{\rot{29.2}}&\multicolumn{1}{c}{\rot{49.7}}&\multicolumn{1}{c}{\rot{32.5}}&\multicolumn{1}{c}{\rot{31.2}}&\multicolumn{1}{c}{\rot{17.8}}&\multicolumn{1}{c}{\rot{5.3}}&\multicolumn{1}{c}{\rot{53.2}}&\multicolumn{1}{c}{\rot{28.8}}&\multicolumn{1}{c}{\rot{36.5}}&\multicolumn{1}{c}{\rot{29.6}}&\multicolumn{1}{c}{\rot{0.0}}&\multicolumn{1}{c}{\rot{14.4}}&\multicolumn{1}{c}{\rot{67.7}}&\multicolumn{1}{c}{\rot{32.4}}&\multicolumn{1}{c}{\rot{10.2}}&\multicolumn{1}{c}{\rot{18.3}}&\multicolumn{1}{c}{\rot{19.2}}&\multicolumn{1}{c}{\rot{11.5}}&\multicolumn{1}{c}{\rot{0.0}}&\multicolumn{1}{c}{\rot{8.9}}&\multicolumn{1}{c}{\rot{38.7}}&\multicolumn{1}{c}{\rot{4.9}}&\multicolumn{1}{c}{\rot{22.6}}&\multicolumn{1}{c}{\rot{55.6}}&\multicolumn{1}{c}{\rot{52.7}}&\multicolumn{1}{c}{\rot{27.9}}&\multicolumn{1}{c}{\rot{29.9}}&\multicolumn{1}{c|}{\rot{8.1}}&{\rot{35.6}}&\multicolumn{1}{c}{\rot{26.3}}\\

      SegNet+DS& \multicolumn{1}{c}{\rot{78.9}}&\multicolumn{1}{c}{\rot{{\bf 92.4}}}&\multicolumn{1}{c}{\rot{52.5}}& \multicolumn{1}{c}{\rot{58.1}}& \multicolumn{1}{c}{\rot{74.1}}& \multicolumn{1}{c}{\rot{{\bf 61.9}}}& \multicolumn{1}{c}{\rot{{\bf 62.6}}}& \multicolumn{1}{c}{\rot{{\bf 64.1}}}& \multicolumn{1}{c}{\rot{60.6}}& \multicolumn{1}{c}{\rot{12.1}}& \multicolumn{1}{c}{\rot{{\bf 70.7}}}& \multicolumn{1}{c}{\rot{{\bf 38.7}}}& \multicolumn{1}{c}{\rot{37.5}}& \multicolumn{1}{c}{\rot{13.1}}& \multicolumn{1}{c}{\rot{{\bf 21.7}}}& \multicolumn{1}{c}{\rot{62.3}}& \multicolumn{1}{c}{\rot{41.2}}& \multicolumn{1}{c}{\rot{55.1}}& \multicolumn{1}{c}{\rot{{\bf 40.1}}}& \multicolumn{1}{c}{\rot{{\bf 25.0}}}&\multicolumn{1}{c}{\rot{{\bf 42.4}}}& \multicolumn{1}{c}{\rot{{\bf 84.0}}}& \multicolumn{1}{c}{\rot{58.1}}& \multicolumn{1}{c}{\rot{{\bf 44.6}}}&\multicolumn{1}{c}{\rot{{\bf 51.4}}}& \multicolumn{1}{c}{\rot{{\bf 44.6}}}& \multicolumn{1}{c}{\rot{{\bf 27.6}}}&\multicolumn{1}{c}{\rot{4.9}}&\multicolumn{1}{c}{\rot{24.8}}&\multicolumn{1}{c}{\rot{{\bf 60.9}}}&\multicolumn{1}{c}{\rot{14.4}}& \multicolumn{1}{c}{\rot{41.1}}&\multicolumn{1}{c}{\rot{{\bf 77.3}}}& \multicolumn{1}{c}{\rot{{\bf 68.7}}}& \multicolumn{1}{c}{\rot{{\bf 60.3}}}& \multicolumn{1}{c}{\rot{63.4}}& \multicolumn{1}{c|}{\rot{{\bf 28.6}}}& {\rot{\textbf{49.2}}}&\multicolumn{1}{c}{\rot{31.2}}\\

       \textbf{LRN}& \multicolumn{1}{c}{\rot{84.6}}&\multicolumn{1}{c}{\rot{90.8}}&\multicolumn{1}{c}{\rot{55.6}}& \multicolumn{1}{c}{\rot{65.2}}&\multicolumn{1}{c}{\rot{{\bf 79.4}}}&\multicolumn{1}{c}{\rot{46.2}}&\multicolumn{1}{c}{\rot{59.8}}&\multicolumn{1}{c}{\rot{44.6}}&\multicolumn{1}{c}{\rot{63.6}}&\multicolumn{1}{c}{\rot{{\bf 32.6}}}& \multicolumn{1}{c}{\rot{66.1}}& \multicolumn{1}{c}{\rot{37.7}}&\multicolumn{1}{c}{\rot{39.8}}&\multicolumn{1}{c}{\rot{{\bf 21.0}}}& \multicolumn{1}{c}{\rot{15.7}}& \multicolumn{1}{c}{\rot{60.2}}& \multicolumn{1}{c}{\rot{{\bf 42.9}}}& \multicolumn{1}{c}{\rot{46.5}}& \multicolumn{1}{c}{\rot{30.4}}& \multicolumn{1}{c}{\rot{20.5}}& \multicolumn{1}{c}{\rot{23.0}}& \multicolumn{1}{c}{\rot{82.8}}& \multicolumn{1}{c}{\rot{{\bf 59.9}}}& \multicolumn{1}{c}{\rot{26.4}}& \multicolumn{1}{c}{\rot{44.4}}& \multicolumn{1}{c}{\rot{32.5}}& \multicolumn{1}{c}{\rot{23.4}}& \multicolumn{1}{c}{\rot{{\bf 10.8}}}& \multicolumn{1}{c}{\rot{{\bf 30.9}}}& \multicolumn{1}{c}{\rot{51.6}}& \multicolumn{1}{c}{\rot{{\bf 21.7}}}& \multicolumn{1}{c}{\rot{{\bf 43.0}}}& \multicolumn{1}{c}{\rot{75.7}}& \multicolumn{1}{c}{\rot{61.8}}& \multicolumn{1}{c}{\rot{51.3}}& \multicolumn{1}{c}{\rot{{\bf 63.9}}}& \multicolumn{1}{c|}{\rot{28.4}}&\rot{46.8}&\multicolumn{1}{c}{\rot{{\bf 33.1}}} \\
       \hline
    \end{tabular}}
\caption{Quantitative results on the SUN RGB-D dataset. We report the accuracy for each class, average class accuracy, and the mean IoU of all classes.}
  \label{tab:quant_cat_sun}
\end{table*}

\subsection{Ablation Analysis}
We perform further ablation analysis to demonstrate the benefit of our coarse-to-fine approach. Our LRN model produces 6 label maps $s_k$ ($k=1,2,...,6$) of different spatial dimensions (see Fig.~\ref{fig:our_archi}). It is possible to obtain the final full-sized label map from any of these $s_k$ by upsamling $s_k$~(using bilinear interpolation) to the spatial dimensions of the input image. Table~\ref{table:stages_pascal} shows the results of this stage-wise analysis on the three datasets. Note that on the PASCAL 2012 dataset, we perform this analysis by learning the model on the training set, and test on the validation set. The reason is that we do not have direct access to the ground-truth labels on the PASCAL 2012 test set. The evaluation on the test set has to be done on the PASCAL evaluation server. The results in Table~\ref{table:stages_pascal} show that if we simply take a label map at one of the early stages of the refinement network and upsample it as the final prediction, the result is not as good as using the label map produced at the last stage. This is reasonable since the label map at an early stage only provides labels corresponding to smaller spatial dimensions.

\begin{table*}[ht!]
\scriptsize
\begin{center}
\resizebox{0.85\textwidth}{!}{
\begin{tabular}{r|c}
 Stage & Mean IoU (\%)\\
\hline\hline
$s_1$ &58.4\\
$s_2$ &61.6\\
$s_3$ &61.9\\
$s_4$ &62.6\\
$s_5$ &62.5\\
$s_6$ &\textbf{62.8}\\
\hline
\multicolumn{2}{c}{(a) PASCAL VOC 2012} \Tstrut \\
\end{tabular}
\hspace{1cm}
\begin{tabular}{r|c}
 Stage & Mean IoU (\%)\\
\hline\hline
$s_1$ &50.9\\
$s_2$ &55.5\\
$s_3$ &59.1\\
$s_4$ &60.9\\
$s_5$ &61.4\\
$s_6$ &\textbf{61.7}\\
\hline
\multicolumn{2}{c}{(b) CamVid} \Tstrut \\
\end{tabular}
\hspace{1cm}
\begin{tabular}{r|c}
 Stage & Mean IoU (\%)\\
\hline\hline
$s_1$ &31.0\\
$s_2$ &31.6\\
$s_3$ &32.4\\
$s_4$ &32.7\\
$s_5$ &32.8\\
$s_6$ &\textbf{33.1}\\ 
\hline
\multicolumn{2}{c}{(c) SUN RGB-D}\Tstrut \\ 
\end{tabular}}
\caption{Stage-wise mean IoU on (a) PASCAL VOC 2012 validation set; (b) CamVid dataset; (c) SUN RGB-D dataset. We can see that, from $s_1$ to $s_6$, mean IoU is progressively enhanced in all the dataset. Note that $s_1..s_6$ are not results of separate training. They are obtained from different stage of the decoder in our model. The result from $s_1$ is implicitly affected by the supervisions provided at $s_2..s_6$, so in Camvid and SUN-RGBD dataset it is reasonable for $s_1$ to produce better performance than SegNet \cite{badrinarayanan15_arxiv}.}
\label{table:stages_pascal}
\end{center}
\end{table*}

\subsection{Discussion}
\begin{figure*}[ht!]
\setlength\tabcolsep{1pt}
\resizebox{\textwidth}{!}{
 \begin{tabular}{ccc}
   \includegraphics[width=0.2\textwidth,height=3cm]{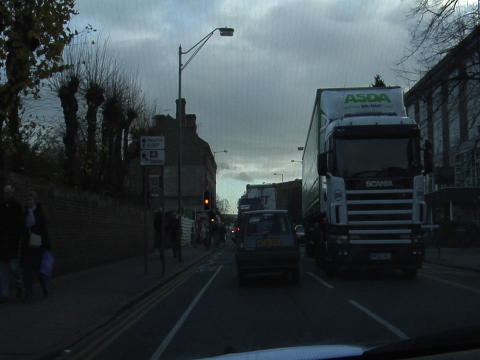}
   &
  \includegraphics[width=0.2\textwidth,height=3cm]{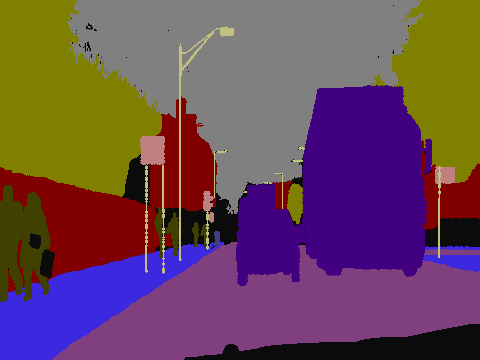}
  &
  {\includegraphics[width=0.2\textwidth,height=3cm]{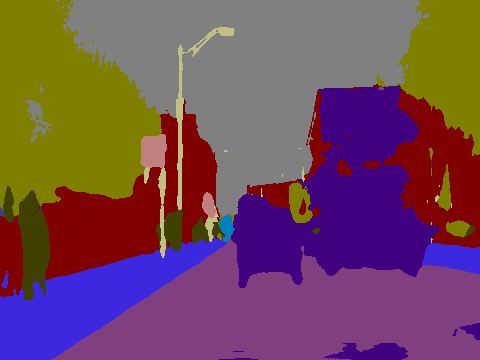}}\\
  \multicolumn{3}{c}{(a)}\\
  \end{tabular}
  \hspace{0.1cm}
  \begin{tabular}{ccc}
  \includegraphics[width=0.2\textwidth,height=3cm]{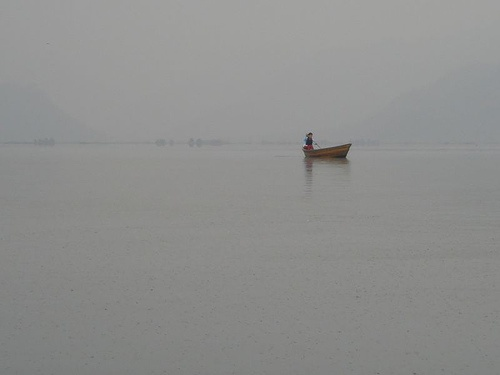}
   &
  \includegraphics[width=0.2\textwidth,height=3cm]{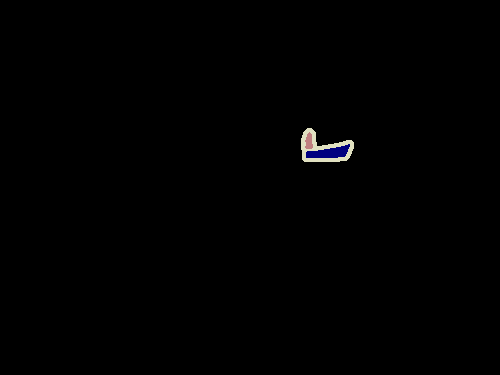}
  &
  {\includegraphics[width=0.2\textwidth,height=3cm]{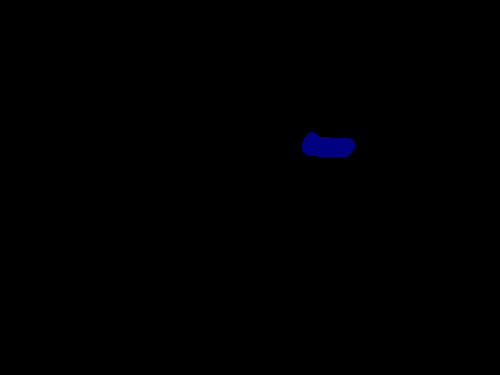}}\\
  \multicolumn{3}{c}{(b)}\\
   \end{tabular}
   \hspace{0.1cm}

   \begin{tabular}{ccc}
   \includegraphics[width=0.2\textwidth,height=3cm]{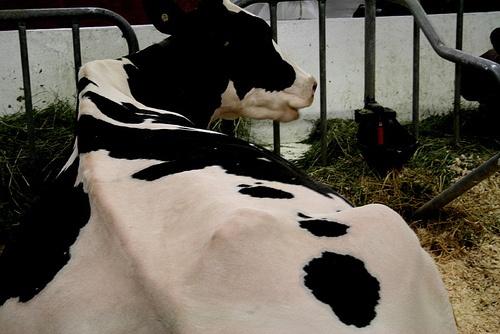}
   &
  \includegraphics[width=0.2\textwidth,height=3cm]{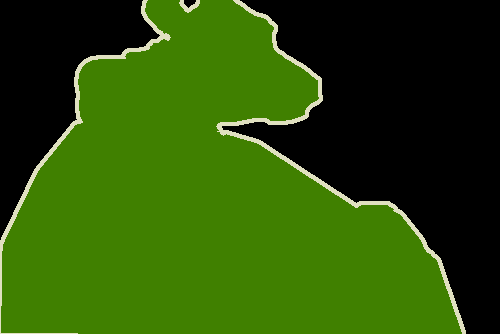}
  &
  {\includegraphics[width=0.2\textwidth,height=3cm]{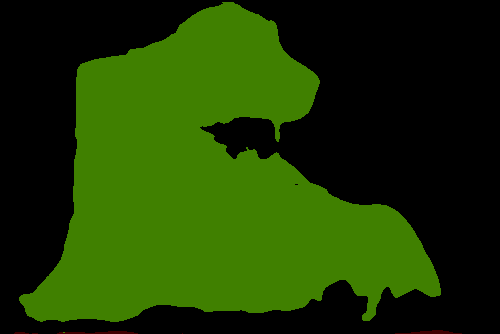}}\\
  \multicolumn{3}{c}{(c)}\\
 \end{tabular}}
\caption{Qualitative analysis of dense predictions on a few problematic images of CamVid \cite{brostow09_prl} and Pascal VOC 2012 \cite{everingham10_ijcv} datasets. Our predictions are more refined and sensitive to smaller and thinner objects (see (a) and (b)). Performance is also better for larger objects (see (c)) because of the refinement modules and guidance driven by early constraints provided by deep supervision.}
  \label{fig:quanlity_analysis}
\end{figure*}

The main focus of our paper is the coarse-to-fine label refinement network architecture. Smaller objects are hard to classify due to the limitation of having a fixed size kernel window. For instance, the poles and sign-symbols in the CamVid dataset \cite{brostow09_prl} often carry significant importance and are important to scene understanding for many applications. This is where our model performs especially well considering the baseline \cite{badrinarayanan15_arxiv}. Table \ref{table:quant_camvid} shows our improvement for classifying column-pole or sign symbols. Figure \ref{fig:quanlity_analysis} shows visually that the prediction of finer details such as the column pole are improved. This improvement is mainly due to the recurrent connection of the weights in the encoder and decoder networks respectively. In addition, the weighted summation of pixels through skip connections help significantly to recover details of smaller objects even after lowering the resolution through several pooling layers.


%% file: conclude.tex
\section{Conclusion}\label{conclude}
We have proposed a novel CNN architecture for semantic segmentation. Our model produces segmentation labels in a coarse-to-fine manner. The segmentation labels at coarse levels are used to progressively refine the labeling produced at finer levels. We introduce multiple loss in the network to provide deep supervisions at multiple stages of the network. Our experimental results demonstrate that our proposed network improves the performance of several baseline approaches. Although we focus on semantic segmentation, the proposed architecture is quite general and can be applied for other pixel-wise labeling problems. In addition, experimental results based on ablation analysis suggest generality in the value of coarse-to-fine predictions, deep supervisions and skip connections respectively, with the implication that a wide range of CNNs may benefit from these architectural modifications.